\documentclass[preprint,12pt]{elsarticle}

\usepackage{amssymb}
\usepackage{amsmath}
\usepackage{graphicx}
\usepackage{amsmath}
\usepackage{amssymb}
\usepackage{amsfonts}
\usepackage{hyperref}
\usepackage{xcolor}
\usepackage{algorithm}
\usepackage{algpseudocode}
\usepackage{lineno}
\usepackage{booktabs}
\usepackage{multirow}
\usepackage{subcaption}

\journal{Nuclear Physics B}

\begin{document}

\begin{frontmatter}



\title{Multi-modal wound classification using wound image and location by Xception and Gaussian Mixture Recurrent Neural Network (GMRNN)}


\author[label1]{Ramin Mousa}
\author[label2]{Ehsan Matbooe}
\author[label1]{Hakimeh Khojasteh}
\author[label3]{Amirali Bengari}
\author[label4]{Mohammadmahdi Vahediahmar}

\affiliation[label1]{organization={Department of Computer Engineering, University of Zanjan},
            addressline={University Blvd},
            state={Zanjan},
            country={Iran}}

\affiliation[label2]{organization={Department of Mathematics, Ferdowsi University of Mashhad},
            addressline={Azadi Square},
            city={Mashhad},
            state={Razavi Khorasan},
            country={Iran}}

\affiliation[label3]{organization={Department of Electrical Engineering, University of Tehran},
            city={Tehran},
            country={Iran}}

\affiliation[label4]{organization={College of Computing and Informatics, Drexel University},
            addressline={3675 Market St},
            city={Philadelphia},
            postcode={19104},
            state={PA},
            country={USA}}

\begin{abstract}
The effective diagnosis of acute and hard-to-heal wounds is crucial for wound care practitioners to provide effective patient care. Poor clinical outcomes are often linked to infection, peripheral vascular disease, and increasing wound depth, which collectively exacerbate these comorbidities. However, diagnostic tools based on Artificial Intelligence (AI) speed up the interpretation of medical images and improve early detection of disease. In this article, we propose a multi-modal AI model based on transfer learning (TL), which combines two state-of-the-art architectures, Xception and GMRNN, for wound classification. The multi-modal network is developed by concatenating the features extracted by a transfer learning algorithm and location features to classify the wound types of diabetic, pressure, surgical, and venous ulcers. The proposed method is comprehensively compared with deep neural networks (DNN) for medical image analysis. The experimental results demonstrate a notable wound-class classifications (containing only diabetic, pressure, surgical, and venous) vary from 78.77 to 100\% in various experiments. The results presented in this study showcase the exceptional accuracy of the proposed methodology in accurately classifying the most commonly occurring wound types using wound images and their corresponding locations.
\end{abstract}



\begin{keyword}
Multi-modal Artificial Intelligence (AI), Wound image classification, Wound location, Gaussian Mixture Recurrent Neural Network(GMRNN), Transfer learning(TL), Recurrent neural network(RNN)



\end{keyword}

\end{frontmatter}



\section{Introduction}
\label{sec:introduction}
Developing diagnostic methods for early detection in the medical field is crucial for providing better treatments and achieving effective outcomes. Among noticeable disruptions, chronic wounds are categorized as hard-to-heal and require early diagnosis and treatment as they affect at least 1.51 to 2.21 per 1000 population\cite{1.}\cite{2.}. Chronic wounds can lead to various complications and increased healthcare costs. With an aging population, the ongoing threat of diabetes and obesity, and persistent infection problems, chronic wounds are expected to remain a significant clinical, social, and economic challenge \cite{3.}\cite{4.}\cite{5.}.
Chronic wound healing is an intricate time-consuming process (healing time 12 weeks). An acute wound is a faster healing wound, whereas, a chronic wound is time-consuming and its healing process is naturally more complicated than an acute wound. The most common types of wounds and ulcers include diabetic foot ulcers (DFUs), venous leg ulcers (VLUs), pressure ulcers (PUs), and surgical wounds (SWs), each involving a significant portion of the population \cite{6.}\cite{7.}. Explainable Artificial Intelligence (XAI) has promisingly been applied in medical research to deliver individualized and data-driven outcomes in wound care. Therefore, use of AI in chronic wound classification appears to be one of the significant keys to serving better treatments \cite{8.}\cite{9.}\cite{10.}.
The tremendous success of AI algorithms in medical image analysis in recent years intersects with a time of dramatically increased use of electronic medical records and diagnostic imaging. Wound diagnosis methods are categorized into machine learning (ML) and deep learning (DL) methods as shown in Figure \ref{Fig1}. Various methods based on machine learning and deep learning, have been developed for wound classification by integrating image and location analysis for wound classification. ML models are designed with explicit features extracted from the input image data. Deep learning models utilize neural networks composed of multiple layers, known as deep neural networks. These networks can learn hierarchical representations of data, enabling them to automatically extract features from raw inputs. This can be advantageous for complex data like medical images or text (e.g., patient records), where feature extraction can be challenging.
Wannous et al. \cite{11.} performed a tissue classification by combining color and texture descriptors as an input vector of an support vector machine (SVM) classifier. They developed a 3D color imaging method for measuring surface area and volume and classifying wound tissues (e.g., granulation, slough, necrosis) to present a single-view and a multi-view approach. Wang et al. \cite{12.}\cite{13.} proposed an approach, using SVM to determine the wound boundaries on foot ulcer images captured with an image capture box. They utilized cascaded two-stage support vector classification to ascertain the DFU region, followed by a two-stage super-pixel classification technique for segmentation and feature extraction.
A machine learning approach was developed by Nagata et al. \cite{14.} to classify skin tears based on the Skin Tear Audit Research (STAR) classification system using digital images, introducing shape features for enhanced accuracy. It compares the performance of support vector machines and random forest algorithms in classifying wound segments and STAR categories. An automated method was proposed by Chitra et al. \cite{15.} for chronic wound tissue classification using the Random Forest (RF) algorithm. They integrated 3-D modeling and unsupervised segmentation techniques to improve accuracy in identifying tissue types such as granulation, slough, and necrotic tissue, achieving a classification accuracy of 93.8\%. Murinto and Sunardi \cite{16.} also evaluated the effectiveness of the SVM algorithm for classifying external wound images. In this research, a feature extraction technique known as the Gray Level Co-occurrence Matrix (GLCM) was employed. GLCM is an image texture analysis method that characterizes the relationship between two adjacent pixels based on their intensity, distance, and grayscale angle.
Sarp et al. \cite{17.} proposed a model for classifying chronic wounds that utilize transfer learning and fully connected layers. Their goal was to improve the interpretability and transparency of AI models, helping clinicians better understand AI-driven diagnoses. The model effectively used transfer learning with VGG16 for feature extraction. Anisuzzaman et al. \cite{18.} presented a multi-modal wound classifier (WMC) network that combines wound images and their corresponding locations to classify different types of wounds. More recently, Mousa et al. \cite{mousa2025integrating} proposed a transformer-based multimodal framework that integrates Vision Transformers and anatomical location data using wavelet augmentation and attention mechanisms, achieving competitive accuracy on the AZH dataset. Utilizing datasets like AZH and Medetec, the study employs a novel deep learning architecture with parallel squeeze-and-excitation blocks, adaptive gated MLP, axial attention mechanism, and convolutional layers. An AI-based system \cite{20.} was developed based on Fast R-CNN and transfer learning techniques for classifying and evaluating diabetic foot ulcers. Fast R-CNN was used for object detection and segmentation. It identifies regions of interest (ROIs) within an image and classifies these regions, while also providing bounding box coordinates for object localization. The model leverages pre-trained convolutional neural networks (CNNs) to improve the learning process on a relatively smaller dataset of diabetic foot wound images.
Scebba et al. \cite{20.} introduced a deep-learning method for automating the segmentation of chronic wound images. This approach employs neural networks to identify and separate wound regions from background noise in the images. The method significantly enhances segmentation accuracy, generalizes effectively to various wound types, and minimizes the need for extensive training data. Another study \cite{21.} combined segmentation with a CNN architecture and a binary classification with traditional ML algorithms to predict surgical site infections in cardiothoracic surgery patients. The system utilizes a MobileNet-Unet model for segmentation and different machine learning classifiers (random forest, support vector machine, and k-nearest neighbors) for classifying wound alterations based on wound type (chest, drain, and leg). Another model based on a convolutional neural network (CNN) \cite{22.} was presented for five wound classification tasks. This model first carries out a phase of feature extraction from the original input image to extract features such as shapes and texture. All extracted features are considered higher-level features, providing semantic information used to classify the input image.
Changa et al. \cite{23.} released a system utilizing multiple deep learning models for automatic burn wound assessment, focusing on accurately estimating the percentage of total body surface area (\%TBSA) burned and segmentation of deep burn regions. The study trained models like U-Net, PSPNet, DeeplabV3+, and Mask R-CNN using boundary-based and region-based labeling methods, achieving high precision and recall. A web-based server was developed to provide automatic burn wound diagnoses and calculate necessary clinical parameters. Another approach was presented by Liu et al. \cite{24.} for automatic segmentation and measurement of pressure injuries using deep learning models and a LiDAR camera. The authors utilized U-Net and Mask R-CNN models to segment wounds from clinical photos and measured wound areas using LiDAR. U-Net outperformed Mask R-CNN in both segmentation and area measurement accuracy. The proposed system achieved acceptable accuracy, showing potential for clinical application in remote monitoring and treatment of pressure injuries.
An XAI model \cite{25.} has been developed to analyze vascular wound images from an Asian population. It leverages deep learning models for wound classification, measurement, and segmentation, achieving high accuracy and explainability. The model utilizes SHAP (Shapley Additive ExPlanations) for model interpretability, providing insights into the decision-making process of the AI, which is crucial for clinical acceptance. A multi-modal wound classification network by Patel et al. \cite{26.} has explored integrating wound location data and image data in classifying pressure injuries using deep learning models. The study employs an Adaptive-gated MLP for separate wound location analysis. Performance metrics vary depending on the number and combination of classes and data splits.

Early detection of chronic wounds is vital for improving treatment outcomes. To meet these important goals, we present an innovative model that combines the strengths of the Xception architecture \cite{27.} and the Capsule Net architecture \cite{28.}. This distinctive integration enhances the model's performance by leveraging transfer learning with pre-trained deep CNNs and meticulous hyperparameter tuning. These methods, extensively validated in medical image analysis, deliver superior results compared to models trained from scratch. Overall, our proposed classification model demonstrates remarkable superiority over other deep learning models, excelling in both accuracy, precision, recall, F1 Score, and specificity. We also performed a sensitivity analysis to investigate setting hyperparameters, batch size, and dropout rate.

\section{Methodology}

\begin{figure*}[!ht]
        \centering
            \includegraphics[width=1\textwidth]{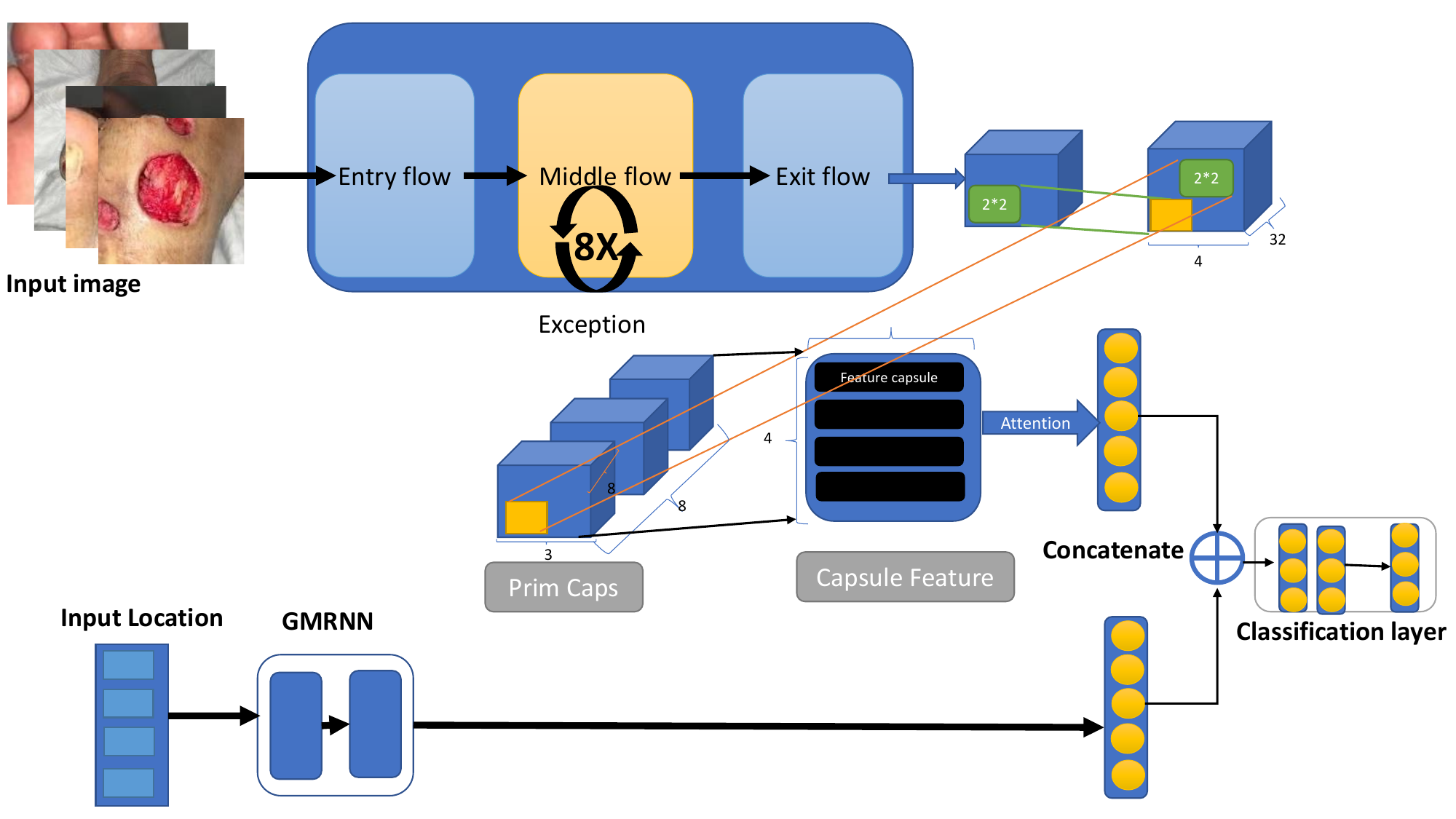}
        \caption{An overview of proposed model.}
        \label{Fig2}
\end{figure*}

An overview of the proposed multimodal classifier network framework is depicted in Figure \ref{Fig2}. Our framework leverages transfer learning for image classification, utilizing the Xception and GMRNN models. Transfer learning helps to develop new machine learning models using pre-trained models from a source task which reduces computational costs. The model employs the Xception architecture as a feature extractor, where robust features are extracted from the images using 2D convolutional layers in Xception. The core concept of Xception lies in its use of depthwise separable convolutions. The Xception model modifies the original Inception block by making it wider and replacing a single $3 * 3$ convolution with a $1 * 1$ convolution to convert the convolution output into low-dimensional embeddings. Then, it performs n spatial transformations, where n denotes the cardinality, indicating the number of transformations and the model's width. This adjustment makes the Xception network more computationally efficient by decoupling spatial and feature-map correlation, which is mathematically indicated in equations (1) and (2):
\begin{equation}\label{d}
  F_{l+1}^K(p,q)=\sum_{x,y}F_l^k(x,y).e_l^k(u,v)
\end{equation}
\begin{equation}\label{d}
  F_{l+2}^k=g_c(F_{l+1}^k,k_{l+1})
\end{equation}
where $k_l$ is a $k$th kernel of the $l$th layer posing depth one, which is spatially convolved across $k$th feature-map $F_l^k$, where $(x,y)$ and $(u,v)$ show the spatial indices of feature-map and kernel respectively. In depthwise separable convolution, it should be noted that the number of kernels $K$ is equal to the number of input feature-maps contrary to a conventional convolutional layer where the number of kernels is independent of previous layer feature-maps. Whereas $k_{l+1}$  is the $k$th kernel of $(1 * 1)$ spatial dimension for $l + 1$th layer, which performs depthwise convolution across output feature-maps $[F_{l+1}^1,…,F_{l+1}^k,…,F_{l+1}^K ]$  of $l$th layer, used as input of $l + 1$th layer.

Xception encoded features are given to the capsule layer. This layer includes a set of capsules. The Capsule covert the scalar features extracted by the Xception layer into vector-valued capsules to capture the input sequence features. If Xception output is $h_i$, and $w$ is a weighted matrix, then $\hat{t}_{i|j}$, which represents the predictor vector, is obtained from the following equation:
      \begin{equation}\label{d}
        \hat{t}_{i|j}=w_{ij}h_i
      \end{equation}
The set of inputs to a capsule $Z_j$ is a weighting set of all prediction vectors $\hat{t}_{i|j}$, which is computed according to the following equation:
      \begin{equation}\label{d}
        Z_j=\sum_{i=1}^Nc_{ij}.\hat{t}_{i|j}
      \end{equation}
Where $c_{ij}$ is the coupling coefficient, which is repeatedly adjusted by Dynamic Routing algorithm [46]. The “squash” is used as a non-linear function for mapping the values of $Z_j$ vectors to [0-1]. This function is applied to $Z_j$  according to the following equation:
      \begin{equation}\label{s}
      v_j=\frac{||Z_j||^2Z_j}{1+||Z_j||^2||Z_j||}
    \end{equation}
The output of a capsule is a vector which can be sent to one of the selected higher-level capsules. In the proposed architecture, Dynamic Routing [46] was used as the routing mechanism.

Self-attention mechanism was applied to select the best and most effective features. Attention is the mapping:
      \begin{equation}\label{d}
        Attention(q, K, V):=\sum_{i=1}^Ksoftmatch_a(q,k)_i.v_i
      \end{equation}
      where $q \in Q$ a query, $Q\subseteq R^d_q$ the query-space, $K\subseteq R^d_k$ the key-space and $K={k_1,...,k_N}\subset K$, $V\subseteq R^d_v$ the value-space and $V={v_1, ..., v_N}\subset V$, and  $softmatch_a(q,k)$ is a probability distribution over the elements of K defined as:
\begin{equation}\label{s}
  softmatch_a(q,K)_i:=\frac{exp(a(q, k_i))}{\sum_{j=1}^{N}}=softmax_j({a(q,k_j)}_j).
\end{equation}
Moreover, when $K=V=Q$ self-attention can be defined as:
\begin{equation}\label{d}
  Q\longmapsto SelfAttention(Q):=Attention(Q, Q, Q).
\end{equation}
The output of Self-attention is the weight vectors obtained from the mapping of the input images. We call this weight vector $Image_{vector}$.

\subsection{Gaussian Mixture Recurrent Neural Network (GMRNN) Cell}

Accounting for uncertainty coefficient, consider linear models and independent samples $y_i$, assume the following distribution for our response $Y$:
\begin{equation}\label{d}
  P(Y|X,\beta)=\Pi_{i=1}^np(y_i,\beta)
\end{equation}
\begin{equation}\label{d}
  p(y_i|x_i,\beta)\sim \aleph(y_i|x_i^T\beta,\sigma^2)
\end{equation}
In addition, since we want to argue about uncertainty coefficient, we also place some distribution $D$ over the parameters $\beta$.
First assume that the coefficient distribution is Gaussian,$\beta\sim\aleph(0,1/\lambda)$, then $D|w\sim\aleph(w^Tx,\sigma^2)$. Using the normal distribution PDF with $\mu$ and $\Sigma$, which in the multivariate case is
\begin{equation}\label{d}
  f(x)=\frac{1}{\sqrt{(2\Pi)^Ndet\Sigma}}exp(-\frac{1}{2}(x-\mu)^T\Sigma(x-\mu))
\end{equation}
$w$ is a normal with $\mu=0$ and $\Sigma=\lambda^{-1}I$ the we get:
\begin{equation}\label{d}
  f(w)=\frac{1}{\sqrt{(2\Pi)^D\frac{1}{\lambda^D}}}exp(-\frac{1}{2}(w-0)^T(\frac{1}{\lambda I})^{-1}(w-0))
\end{equation}
For getting the $f(D|w)$ first need $f(y_k|w)$:
\begin{equation}\label{d}
  f(y_k|w)=\frac{1}{\sqrt{2\Pi\sigma^2}}exp(-\frac{1}{2\sigma^2}(y_k-x^Tw)^2)
\end{equation}
but $y_1,...,y_{D|w}$ are independent, then:
\begin{multline}
  f(D|w)=f(y_1,...,y_{D|w}=\Pi_{k=1}^Nf(y_k|w)=\\ \Pi_{k=1}^N\frac{1}{\sqrt{2\Pi\sigma^2}}exp(-\frac{1}{2\sigma^2}(y_k-x^Tw)^2)
\end{multline}

Now, having the logarithm of the relationship \ref{EED}, which is calculated as $\log P(w|D)=\log P(D|w)+\log P(w)-\log P(D)$ , MAP maximizes  with respect to $w$ is obtained as follows:
\begin{equation}\label{e}
  \hat{w}=argmax_w \log P(w|D)
\end{equation}
That is equal to:
\begin{equation}\label{s}
    \hat{w}=argmax_w(\log P(D|w)+\log P(w)-\log P(D))
\end{equation}
Considering that $P(D)$ is independent of $w$, the following relation can be adapted:

\begin{equation}\label{s}
    \hat{w}=argmax_w(\log P(D|w)+\log P(w))
\end{equation}
where $\log P(D|w)$ calculated as follows:
 \begin{align}
    \log P(D|w) &= \log\left(\prod_{k=1}^D \frac{1}{\sqrt{2\pi\sigma^2}} \exp\left(-\frac{1}{2\sigma^2}(y_k-x^Tw)^2\right)\right) \\
    &= \sum_{k=1}^D \log\frac{1}{\sqrt{2\pi\sigma^2}} - \frac{1}{2\sigma^2}\sum_{k=1}^N(y_k-x^Tw)^2 \\
    &= D\log\frac{1}{\sqrt{2\pi\sigma^2}} - \frac{1}{2\sigma^2}\sum_{k=1}^N(y_k-x^Tw)^2
\end{align}
\\
The log value for $f(w)$ calculate as follow:
\begin{equation}\label{s}
  \log f(w)=\log \lambda^{\frac{D}{2}}-\log(2\pi)^{\frac{D}{2}}-\frac{\lambda}{2}w^Tw
\end{equation}
by having the $\log P(D|w) $ and $\log f(w)$, $\hat{w}$ calculate as follow:
\begin{align}
    \hat{w} &= \arg\max_w \left(D\log\frac{1}{\sqrt{2\pi\sigma^2}} - \frac{1}{2\sigma^2}\sum_{k=1}^N(y_k - x^T w)^2 + \right. \nonumber \\
    &\quad \left. \log \lambda^{\frac{D}{2}} - \log(2\pi)^{\frac{D}{2}} - \frac{\lambda}{2} w^T w \right) \\
    &= \arg\max_w \left(-\frac{1}{2\sigma^2} \sum_{k=1}^N (y_k - x^T w)^2 - \frac{\lambda}{2} w^T w \right)
\end{align}

  Maximizing $-x$is equal to minimizing $x$ if $x\geq0$,hence:

\begin{equation}\label{d}
  argmin_w(\frac{1}{2\sigma^2}\Sigma_{k=1}^N(y_k-x^Tw)^2-\frac{\lambda}{2}w^Tw)
\end{equation}
According to the investigated relationships, GMRNN can be defined as follows:
\begin{equation}\label{w}
  f^{<k>} = \sigma(W_f x^{<k>}+U_f h^{<k-1>}+b_f)
\end{equation}

\begin{equation}\label{w}
  i^{<k>} = \sigma(W_i x^{<k>}+U_i h^{<k-1>}+b_i)
\end{equation}

\begin{equation}\label{w}
  g^{<k>} =  tanh(W_g x^{<k>}+U_g h^{<k-1>}+b_g)
\end{equation}

\begin{equation}\label{w}
o^{<k>} =  \sigma(W_o x^{<k>}+U_o h^{<k-1>}+b_o)
\end{equation}

\begin{equation}\label{w}
m^{<k>} =  \sigma(W_o \hat{w}^{<k>}+U_m h^{<k-1>}+b_m)
\end{equation}

\begin{equation}\label{s}
 C_t= \sigma( f^{<k>}*C_{t-1}+i^{<k>}*g^{<k>} + m^{<k>})
\end{equation}
\begin{equation}\label{s}
  h_t=tanh(C_t)*o^{<k>}
\end{equation}
The output of the GMRNN consists of weight vectors generated from mapping wound locations to the GMRNN. This output layer, represented by the weight vectors, is referred to as $Location_{vector}$.
The integration of two vectors is obtained as follows:
\begin{equation}\label{s}
  output_{vector} = Image_{vector}\bigoplus Locaton_{vector}
\end{equation}

$Output_{vector}$ after passing through several layers is completely connected to a layer with $N$ neurons ($N$ number of classes) for classification. Softmax function is used to calculate the probability of each class.

\section{Evaluation tools}

\subsection{Dataset}
In this study, we used the AZH dataset, provided by Anisuzzaman et al. \cite{18.}. This dataset was collected over a two-year clinical period at the AZH Wound and Vascular Center in Milwaukee, Wisconsin. It consists of 730 wound images in .jpg format and is publicly available in this GitHub\footnote{\href{https://github.com/uwm-bigdata/Multi-modal-wound-classification-using-images-and-locations/tree/main}{GitHub}} repository. The images vary in size, with widths ranging from 320 to 700 pixels and heights ranging from 240 to 525 pixels. The dataset includes four different wound types: venous, diabetic, pressure, and surgical. Most images in the dataset were taken from different patients, but some images were captured from the same patient at different body sites or various stages of healing. Additionally, in cases where the wound shapes differed, they were considered separate images.

\begin{figure*}[!ht]
        \centering
            \includegraphics[ width=1\textwidth,height=13cm]{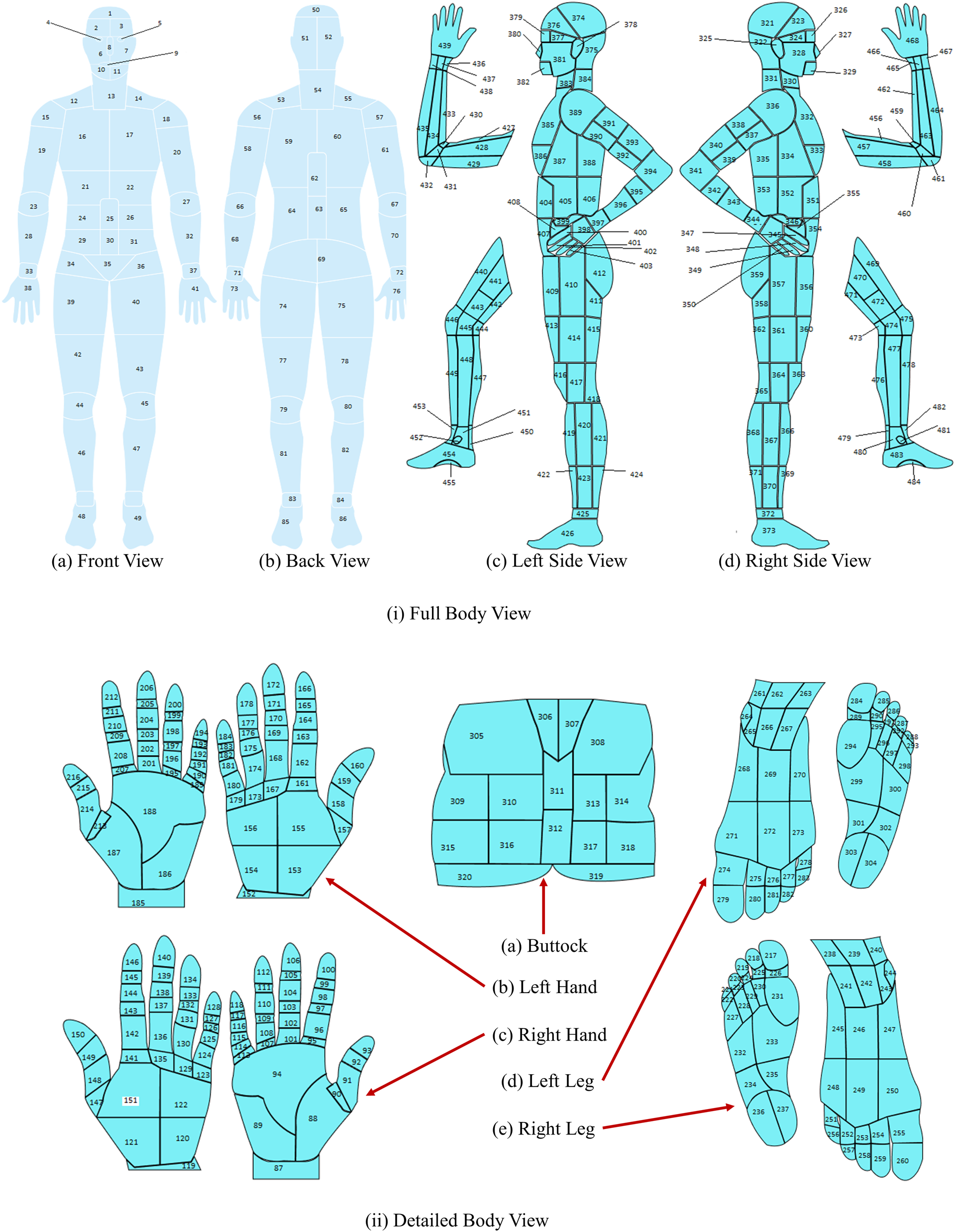}
        \caption{Body map for location selection(image tacked from \cite{BASE}).}
        \label{FigBo}
\end{figure*}

\begin{figure*}[!ht]
        \centering
            \includegraphics[ width=1\textwidth,height=13cm]{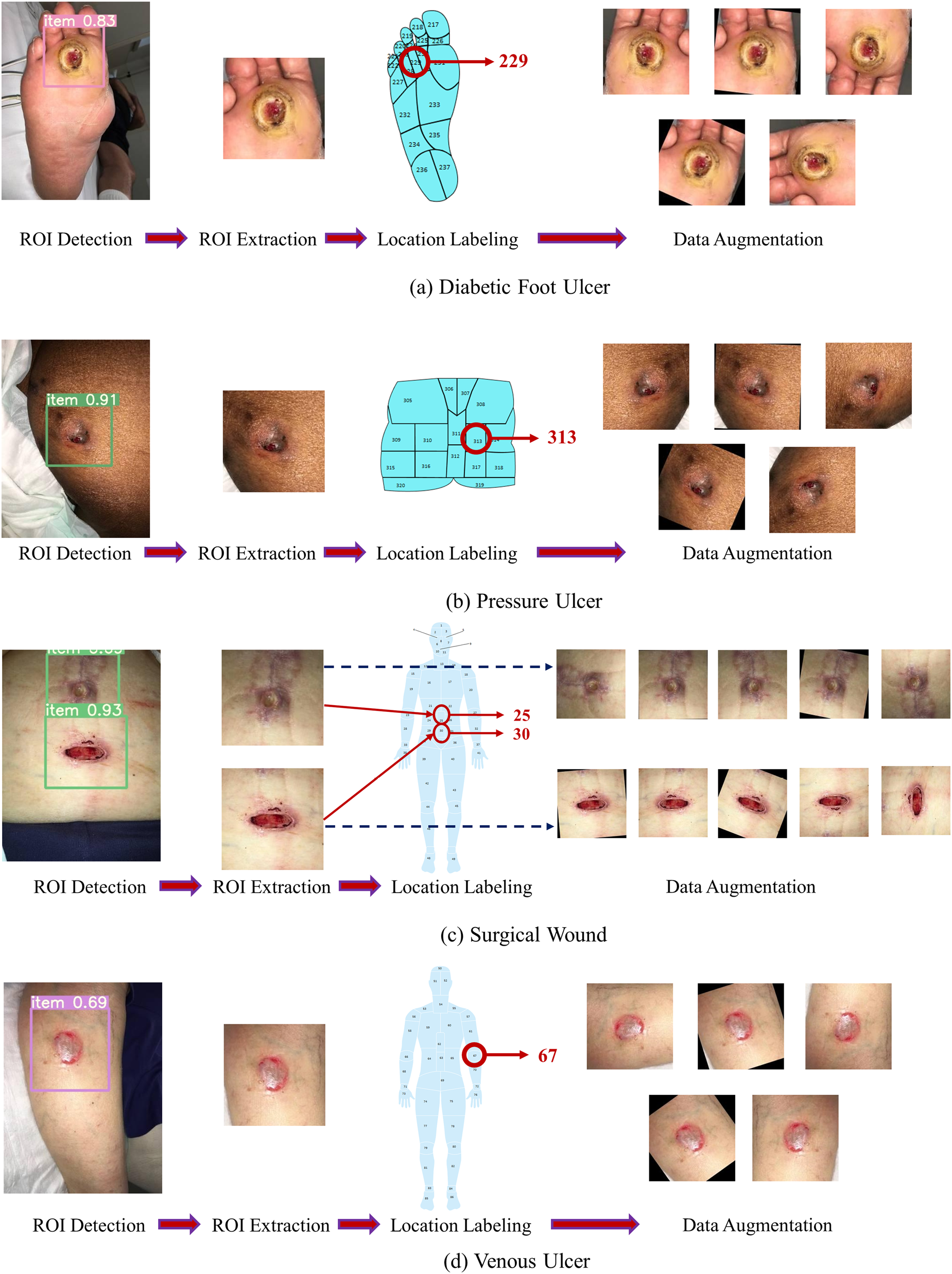}
        \caption{Dataset processing steps(image tacked from \cite{BASE}).}
        \label{FigBo2}
\end{figure*}

\begin{figure*}
\centering
\subfloat[diabetic ]{\includegraphics[width=3cm,height=3cm]{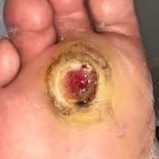}}
\subfloat[]{\includegraphics[ width=3cm,height=3cm]{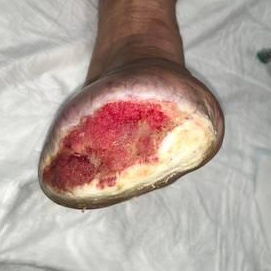}}
\subfloat[]{\includegraphics[width=3cm,height=3cm]{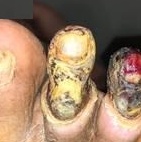}}
\subfloat[]{\includegraphics[width=3cm,height=3cm]{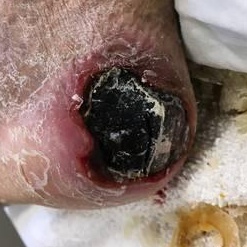}}
\subfloat[]{\includegraphics[width=3cm,height=3cm]{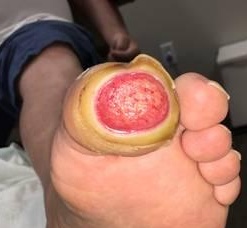}}\\
\subfloat[pressure ]{\includegraphics[width=3cm,height=3cm]{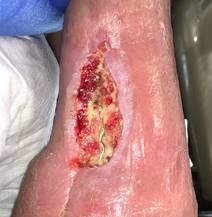}}
\subfloat[]{\includegraphics[ width=3cm,height=3cm]{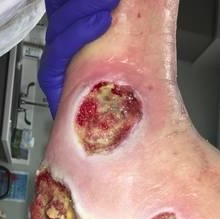}}
\subfloat[]{\includegraphics[width=3cm,height=3cm]{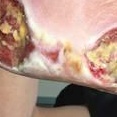}}
\subfloat[]{\includegraphics[width=3cm,height=3cm]{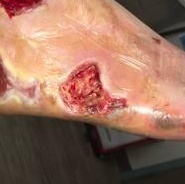}}
\subfloat[]{\includegraphics[width=3cm,height=3cm]{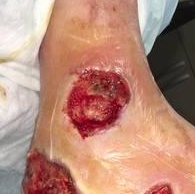}}\\
\subfloat[surgical ]{\includegraphics[width=3cm,height=3cm]{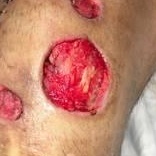}}
\subfloat[]{\includegraphics[ width=3cm,height=3cm]{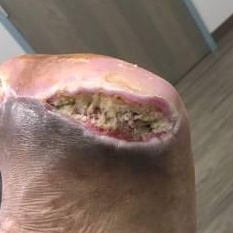}}
\subfloat[]{\includegraphics[width=3cm,height=3cm]{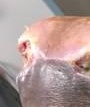}}
\subfloat[]{\includegraphics[width=3cm,height=3cm]{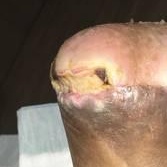}}
\subfloat[]{\includegraphics[width=3cm,height=3cm]{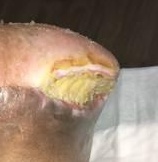}}\\
\subfloat[venous ]{\includegraphics[width=3cm,height=3cm]{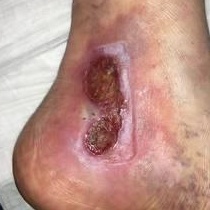}}
\subfloat[]{\includegraphics[ width=3cm,height=3cm]{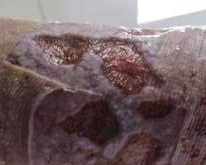}}
\subfloat[]{\includegraphics[width=3cm,height=3cm]{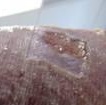}}
\subfloat[]{\includegraphics[width=3cm,height=3cm]{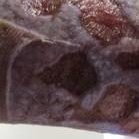}}
\subfloat[]{\includegraphics[width=3cm,height=3cm]{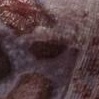}}\\
\caption{Some examples of data sets.}
\label{BarChartSa}
\end{figure*}

In this research, we also utilized the body map developed by Anisuzzaman et al.\cite{BASE}, which demonstrates exactly where the wound is located. A body map is a chart and visual tool primarily used in the healthcare sector to precisely document and track various health conditions. By employing a generic image of a body that roughly represents the target audience, such as a male or female adult or child, one can accurately indicate where the individual is experiencing a health-related issue or receiving treatment \cite{B29}. Body maps can also be used in wound assessment to examine various types of wounds, including abrasions, lacerations, burns, surgical incisions, pressure injuries, skin tears, arterial ulcers, and venous ulcers. Understanding the type of wound is crucial for selecting appropriate interventions. The wound's location should be documented precisely, and a body diagram template is useful for accurately indicating the wound's position. Additionally, the size of the wound should be measured regularly to monitor any changes, determining if the wound is increasing or decreasing in size \cite{B30}.
The released body map by Anisuzzaman et al. \cite{BASE} which includes 484 distinct parts. By using a total of 484 features or regions, they avoided the extreme intricacy of depicting every detailed feature of the body. These regions were pre-selected and validated by wound professionals at the AZH Wound and Vascular Center. The resulting body map is shown in Figure \ref{FigBo}, with each number representing a specific location. During the experiments, they generated a simplified body map by merging various sections of the original due to a lack of images for some wound types and locations. For instance, body locations 436, 437, and 438 were combined and labeled as 436, while body locations 390, 391, 392, and 393 were merged and labeled as 390, and so on. This simplification removed 161 location points, reducing the total number of locations from 484 to 323. Examples of the simplified body map are shown in Figure \ref{FigBo2}. This simplified body map, containing 323 locations, was used in this work.Some examples of data sets are shown in Figure \ref{BarChartSa}.

\subsection{Deep learning library}

Keras\footnote{\url{https://keras.io/}} is a straightforward API; it has a standard interface and behaviours in which the model's components can be easily shared and debugged. The best thing you can say about any software library is that the abstractions it chooses are completely natural, so there is no friction between thinking about what you want to do and how to code it. That is exactly what you get. Keras allows us to prototype, research and deploy deep learning models intuitively and efficiently. The functional API makes the code understandable and lightweight, enabling effective knowledge transfer between team scientists. This API is provided under the backend of Google's TensorFlow\footnote{\url{https://www.tensorflow.org/}}, MILA's Theano\footnote{\url{https://pypi.org/project/Theano}} or Microsoft's CNTK\footnote{\url{https://learn.microsoft.com/en-us/cognitive-toolkit/}}, and Apache's MXNet\footnote{\url{https://mxnet.apache.org/}}.
Our work uses Keras to develop the neural network model described in Section 4. From the implementation point of view, the Keras library is used. Tensorflow was used as the back layer on which the Keras backend runs. Our proposed model uses a combination of different convolutional layers, capsule layers, and encephalic learning, and defining these blocks in Keras is easier.
A Computation Graph Configuration may have any number of inputs (multiple independent inputs, possibly of different types) and any number of output layers. This is the second reason we chose this tool to develop our network. Our model has three distinct input layers.

\subsection{Evaluation Metrics}
We used the following evaluation metrics to assess the performance of our proposed model: accuracy, precision, recall, F1 Score, and specificity.

\begin{equation}\label{s}
  Accuracy=\frac{(TP+TN)}{(TP+TN+FP+FN)}
\end{equation}

\begin{equation}\label{s}
Precision=\frac{TP}{TP+FP}
\end{equation}

\begin{equation}\label{s}
Recall=\frac{TP}{TP+FN}
\end{equation}

\begin{equation}\label{s}
F1-score=2*\frac{precision*recall}{precision+recall}
\end{equation}

\begin{equation}\label{s}
Specificity=\frac{TN}{TN+FP}
\end{equation}

\section{Result}
We conducted several experiments on the AZH dataset, focusing on four-wound class classifications (D vs. P vs. S vs. V), to identify the optimal model combinations for the proposed model. This classification task was the most challenging, as the experiment did not include any normal skin (N) or background (BG) images. The experiment was performed using a custom-developed body map, which comprises 484 locations. Table \ref{Table1} displays the results of these experiments. Furthermore, we present the results on the original dataset (without any augmentation) to demonstrate the impact (improvement) of data augmentation. The performances of MLP and LSTM were similar on the location data, whereas GMRNN was the best with an accuracy of 0.6923. On the original image data, MobileNetV2 + Capsule, Densenet121 + Capsule, VGG16 + Capsule, and InceptionV3 + Capsule achieved almost the same accuracy. We concluded that Capsule was a consistent model to boost model performance. The performances of AlexNet + MLP, AlexNet + LSTM, and ResNet50 + LSTM were poor as shown in Table \ref{Table1}. However, the Xception + Capsule performed best on the image data.  Running all these combinations for multiple experiments was also time-consuming and memory-intensive. Therefore, based on these results, we selected the top five combinations (VGG16 + MLP, VGG19 + MLP, VGG16 + LSTM, VGG19 + LSTM, and Xception + GMRNN) for all subsequent experimental setups.

\begin{table*}[]

\centering
\resizebox{\textwidth}{!}{
\begin{tabular}{|l|l|llllll|lllll|l|}
\hline
                                   &                      & \multicolumn{1}{l|}{Accuracy} & \multicolumn{1}{l|}{Precision} & \multicolumn{1}{l|}{Recall} & \multicolumn{1}{l|}{F1}     & \multicolumn{1}{l|}{specificity} & sensitivity & \multicolumn{1}{l|}{Accuracy} & \multicolumn{1}{l|}{Precision} & \multicolumn{1}{l|}{Recall} & \multicolumn{1}{l|}{F1}     & specificity & sensitivity \\ \hline
\multirow{3}{*}{Location}          &                      & \multicolumn{6}{c|}{Original Data}                                                                                                                                          & \multicolumn{5}{c|}{Augmented data}                                                                                                      &             \\ \cline{2-14}
                                   & MLP                  & \multicolumn{1}{l|}{0.6630}   & \multicolumn{1}{l|}{-}         & \multicolumn{1}{l|}{-}      & \multicolumn{1}{l|}{-}      & \multicolumn{1}{l|}{-}           & -           & \multicolumn{1}{l|}{0.7174}   & \multicolumn{1}{l|}{-}         & \multicolumn{1}{l|}{-}      & \multicolumn{1}{l|}{-}      & -           & -           \\ \cline{2-14}
                                   & LSTM                 & \multicolumn{1}{l|}{0.6685}   & \multicolumn{1}{l|}{-}         & \multicolumn{1}{l|}{-}      & \multicolumn{1}{l|}{-}      & \multicolumn{1}{l|}{-}           & -           & \multicolumn{1}{l|}{0.7228}   & \multicolumn{1}{l|}{-}         & \multicolumn{1}{l|}{-}      & \multicolumn{1}{l|}{-}      & -           & -           \\ \cline{2-14}
                                   & GMRNN                & \multicolumn{1}{l|}{0.6923}   & \multicolumn{1}{l|}{0.7014}    & \multicolumn{1}{l|}{0.6988} & \multicolumn{1}{l|}{0.7001} & \multicolumn{1}{l|}{0.9709}      & 0.8162      & \multicolumn{1}{l|}{0.7479}   & \multicolumn{1}{l|}{0.7473}    & \multicolumn{1}{l|}{0.7449} & \multicolumn{1}{l|}{0.7461} & 0.9735      & 0.8462      \\ \hline
\multirow{13}{*}{Image}            & AlexNet              & \multicolumn{1}{l|}{0.3533}   & \multicolumn{1}{l|}{-}         & \multicolumn{1}{l|}{-}      & \multicolumn{1}{l|}{-}      & \multicolumn{1}{l|}{-}           & -           & \multicolumn{1}{l|}{0.3750}   & \multicolumn{1}{l|}{-}         & \multicolumn{1}{l|}{-}      & \multicolumn{1}{l|}{-}      & -           & -           \\ \cline{2-14}
                                   & VGG16                & \multicolumn{1}{l|}{0.6576}   & \multicolumn{1}{l|}{-}         & \multicolumn{1}{l|}{-}      & \multicolumn{1}{l|}{-}      & \multicolumn{1}{l|}{-}           & -           & \multicolumn{1}{l|}{0.7173}   & \multicolumn{1}{l|}{-}         & \multicolumn{1}{l|}{-}      & \multicolumn{1}{l|}{-}      & -           & -           \\ \cline{2-14}
                                   & VGG19                & \multicolumn{1}{l|}{0.5652}   & \multicolumn{1}{l|}{-}         & \multicolumn{1}{l|}{-}      & \multicolumn{1}{l|}{-}      & \multicolumn{1}{l|}{-}           & -           & \multicolumn{1}{l|}{0.6304}   & \multicolumn{1}{l|}{-}         & \multicolumn{1}{l|}{-}      & \multicolumn{1}{l|}{-}      & -           & -           \\ \cline{2-14}
                                   & InceptionV3          & \multicolumn{1}{l|}{0.5109}   & \multicolumn{1}{l|}{-}         & \multicolumn{1}{l|}{-}      & \multicolumn{1}{l|}{-}      & \multicolumn{1}{l|}{-}           & -           & \multicolumn{1}{l|}{0.5609}   & \multicolumn{1}{l|}{-}         & \multicolumn{1}{l|}{-}      & \multicolumn{1}{l|}{-}      & -           & -           \\ \cline{2-14}
                                   & ResNet50             & \multicolumn{1}{l|}{0.3370}   & \multicolumn{1}{l|}{-}         & \multicolumn{1}{l|}{-}      & \multicolumn{1}{l|}{-}      & \multicolumn{1}{l|}{-}           & -           & \multicolumn{1}{l|}{0.3370}   & \multicolumn{1}{l|}{-}         & \multicolumn{1}{l|}{-}      & \multicolumn{1}{l|}{-}      & -           & -           \\ \cline{2-14}
                                   & MobileNetV2 + Capsule           & \multicolumn{1}{l|}{0.6771}   & \multicolumn{1}{l|}{0.6667}    & \multicolumn{1}{l|}{0.6667} & \multicolumn{1}{l|}{0.6667} & \multicolumn{1}{l|}{0.9604}      & 0.9271      & \multicolumn{1}{l|}{0.7420}   & \multicolumn{1}{l|}{0.7470}    & \multicolumn{1}{l|}{0.789}  & \multicolumn{1}{l|}{0.7674} & 0.9735      & 0.8462      \\ \cline{2-14}
                                   & Densenet121 + Capsule           & \multicolumn{1}{l|}{0.6771}   & \multicolumn{1}{l|}{0.6756}    & \multicolumn{1}{l|}{0.6641} & \multicolumn{1}{l|}{0.6698} & \multicolumn{1}{l|}{0.9677}      & 0.8229      & \multicolumn{1}{l|}{0.6413}   & \multicolumn{1}{l|}{0.6568}    & \multicolumn{1}{l|}{0.6297} & \multicolumn{1}{l|}{0.6429} & 0.9203      & 0.8913      \\ \cline{2-14}
                                   & VGG16  + Capsule                & \multicolumn{1}{l|}{0.6771}   & \multicolumn{1}{l|}{0.6667}    & \multicolumn{1}{l|}{0.6667} & \multicolumn{1}{l|}{0.6667} & \multicolumn{1}{l|}{0.9646}      & 0.9427      & \multicolumn{1}{l|}{0.7290}   & \multicolumn{1}{l|}{0.7212}    & \multicolumn{1}{l|}{0.6757} & \multicolumn{1}{l|}{0.6977} & 0.9312      & 0.9587      \\ \cline{2-14}
                                   & VGG19  + Capsule                & \multicolumn{1}{l|}{0.6510}   & \multicolumn{1}{l|}{0.6436}    & \multicolumn{1}{l|}{0.6410} & \multicolumn{1}{l|}{0.6422} & \multicolumn{1}{l|}{0.9656}      & 0.9583      & \multicolumn{1}{l|}{0.7173}   & \multicolumn{1}{l|}{0.7145}    & \multicolumn{1}{l|}{0.7123} & \multicolumn{1}{l|}{0.7133} & 0.9312      & 0.8462      \\ \cline{2-14}
                                   & Xception + Capsule              & \multicolumn{1}{l|}{0.6875}   & \multicolumn{1}{l|}{0.6885}    & \multicolumn{1}{l|}{0.6795} & \multicolumn{1}{l|}{0.6839} & \multicolumn{1}{l|}{0.9552}      & 0.7500      & \multicolumn{1}{l|}{0.7589}   & \multicolumn{1}{l|}{0.7799}    & \multicolumn{1}{l|}{0.7569} & \multicolumn{1}{l|}{0.7682} & 0.9838      & 0.8932      \\ \cline{2-14}
                                   & InceptionV3 + Capsule           & \multicolumn{1}{l|}{0.6719}   & \multicolumn{1}{l|}{0.6731}    & \multicolumn{1}{l|}{0.6564} & \multicolumn{1}{l|}{0.6646} & \multicolumn{1}{l|}{0.9479}      & 0.9271      & \multicolumn{1}{l|}{0.6838}   & \multicolumn{1}{l|}{0.6778}    & \multicolumn{1}{l|}{0.6562} & \multicolumn{1}{l|}{0.6667} & 0.9598      & 0.8735      \\ \cline{2-14}
                                   & EfficientNetB0 + Capsule        & \multicolumn{1}{l|}{0.5729}   & \multicolumn{1}{l|}{0.5701}    & \multicolumn{1}{l|}{0.5026} & \multicolumn{1}{l|}{0.5342} & \multicolumn{1}{l|}{0.9323}      & 0.8906      & \multicolumn{1}{l|}{0.7271}   & \multicolumn{1}{l|}{0.7473}    & \multicolumn{1}{l|}{0.7449} & \multicolumn{1}{l|}{0.7461} & 0.9735      & 0.8462      \\ \cline{2-14}
                                   & ResNet50 + Capsule             & \multicolumn{1}{l|}{0.6823}   & \multicolumn{1}{l|}{0.6769}    & \multicolumn{1}{l|}{0.6667} & \multicolumn{1}{l|}{0.6667} & \multicolumn{1}{l|}{0.9667}      & 0.9583      & \multicolumn{1}{l|}{0.6196}   & \multicolumn{1}{l|}{0.6212}    & \multicolumn{1}{l|}{0.5757} & \multicolumn{1}{l|}{0.5975} & 0.9112      & 0.8587      \\ \hline
\multirow{11}{*}{Image + Location} & AlexNet +   MLP      & \multicolumn{1}{l|}{0.5543}   & \multicolumn{1}{l|}{-}         & \multicolumn{1}{l|}{-}      & \multicolumn{1}{l|}{-}      & \multicolumn{1}{l|}{-}           & -           & \multicolumn{1}{l|}{0.6141}   & \multicolumn{1}{l|}{-}         & \multicolumn{1}{l|}{-}      & \multicolumn{1}{l|}{-}      & -           & -           \\ \cline{2-14}
                                   & VGG16 + MLP          & \multicolumn{1}{l|}{0.7717}   & \multicolumn{1}{l|}{-}         & \multicolumn{1}{l|}{-}      & \multicolumn{1}{l|}{-}      & \multicolumn{1}{l|}{-}           & -           & \multicolumn{1}{l|}{0.78}     & \multicolumn{1}{l|}{-}         & \multicolumn{1}{l|}{-}      & \multicolumn{1}{l|}{-}      & -           & -           \\ \cline{2-14}
                                   & VGG19 + MLP          & \multicolumn{1}{l|}{0.6250}   & \multicolumn{1}{l|}{-}         & \multicolumn{1}{l|}{-}      & \multicolumn{1}{l|}{-}      & \multicolumn{1}{l|}{-}           & -           & \multicolumn{1}{l|}{0.7228}   & \multicolumn{1}{l|}{-}         & \multicolumn{1}{l|}{-}      & \multicolumn{1}{l|}{-}      & -           & -           \\ \cline{2-14}
                                   & InceptionV3   + MLP  & \multicolumn{1}{l|}{0.6141}   & \multicolumn{1}{l|}{-}         & \multicolumn{1}{l|}{-}      & \multicolumn{1}{l|}{-}      & \multicolumn{1}{l|}{-}           & -           & \multicolumn{1}{l|}{0.711}    & \multicolumn{1}{l|}{-}         & \multicolumn{1}{l|}{-}      & \multicolumn{1}{l|}{-}      & -           & -           \\ \cline{2-14}
                                   & ResNet50 +   MLP     & \multicolumn{1}{l|}{0.6304}   & \multicolumn{1}{l|}{-}         & \multicolumn{1}{l|}{-}      & \multicolumn{1}{l|}{-}      & \multicolumn{1}{l|}{-}           & -           & \multicolumn{1}{l|}{0.6685}   & \multicolumn{1}{l|}{-}         & \multicolumn{1}{l|}{-}      & \multicolumn{1}{l|}{-}      & -           & -           \\ \cline{2-14}
                                   & AlexNet +   LSTM     & \multicolumn{1}{l|}{0.5815}   & \multicolumn{1}{l|}{-}         & \multicolumn{1}{l|}{-}      & \multicolumn{1}{l|}{-}      & \multicolumn{1}{l|}{-}           & -           & \multicolumn{1}{l|}{0.6685}   & \multicolumn{1}{l|}{-}         & \multicolumn{1}{l|}{-}      & \multicolumn{1}{l|}{-}      & -           & -           \\ \cline{2-14}
                                   & VGG16 +   LSTM       & \multicolumn{1}{l|}{0.7283}   & \multicolumn{1}{l|}{-}         & \multicolumn{1}{l|}{-}      & \multicolumn{1}{l|}{-}      & \multicolumn{1}{l|}{-}           & -           & \multicolumn{1}{l|}{0.7935}   & \multicolumn{1}{l|}{-}         & \multicolumn{1}{l|}{-}      & \multicolumn{1}{l|}{-}      & -           & -           \\ \cline{2-14}
                                   & VGG19 +   LSTM       & \multicolumn{1}{l|}{0.71200}  & \multicolumn{1}{l|}{-}         & \multicolumn{1}{l|}{-}      & \multicolumn{1}{l|}{-}      & \multicolumn{1}{l|}{-}           & -           & \multicolumn{1}{l|}{0.7663}   & \multicolumn{1}{l|}{-}         & \multicolumn{1}{l|}{-}      & \multicolumn{1}{l|}{-}      & -           & -           \\ \cline{2-14}
                                   & InceptionV3   + LSTM & \multicolumn{1}{l|}{0.6467}   & \multicolumn{1}{l|}{-}         & \multicolumn{1}{l|}{-}      & \multicolumn{1}{l|}{-}      & \multicolumn{1}{l|}{-}           & -           & \multicolumn{1}{l|}{0.692}    & \multicolumn{1}{l|}{-}         & \multicolumn{1}{l|}{-}      & \multicolumn{1}{l|}{-}      & -           & -           \\ \cline{2-14}
                                   & ResNet50 +   LSTM    & \multicolumn{1}{l|}{0.3370}   & \multicolumn{1}{l|}{-}         & \multicolumn{1}{l|}{-}      & \multicolumn{1}{l|}{-}      & \multicolumn{1}{l|}{-}           & -           & \multicolumn{1}{l|}{0.3479}   & \multicolumn{1}{l|}{-}         & \multicolumn{1}{l|}{-}      & \multicolumn{1}{l|}{-}      & -           & -           \\ \cline{2-14}
                                   & Xception+ GMRNN      & \multicolumn{1}{l|}{0.7877}   & \multicolumn{1}{l|}{0.7882}    & \multicolumn{1}{l|}{0.7715} & \multicolumn{1}{l|}{0.7797} & \multicolumn{1}{l|}{0.9662}      & 0.9334      & \multicolumn{1}{l|}{0.8189}   & \multicolumn{1}{l|}{0.8159}    & \multicolumn{1}{l|}{0.8469} & \multicolumn{1}{l|}{0.8311} & 0.9865      & 0.8944      \\ \hline
\end{tabular}
}
\caption{Four wound class classification (D vs. P vs. S vs. V) on AZH dataset with original body map. The bold
represents the highest results/accuracy achieved for each experiment.}
\label{Table1}
\end{table*}

\begin{figure*}
\centering
\subfloat[Original Data Location]{\includegraphics[width=5cm,height=5cm]{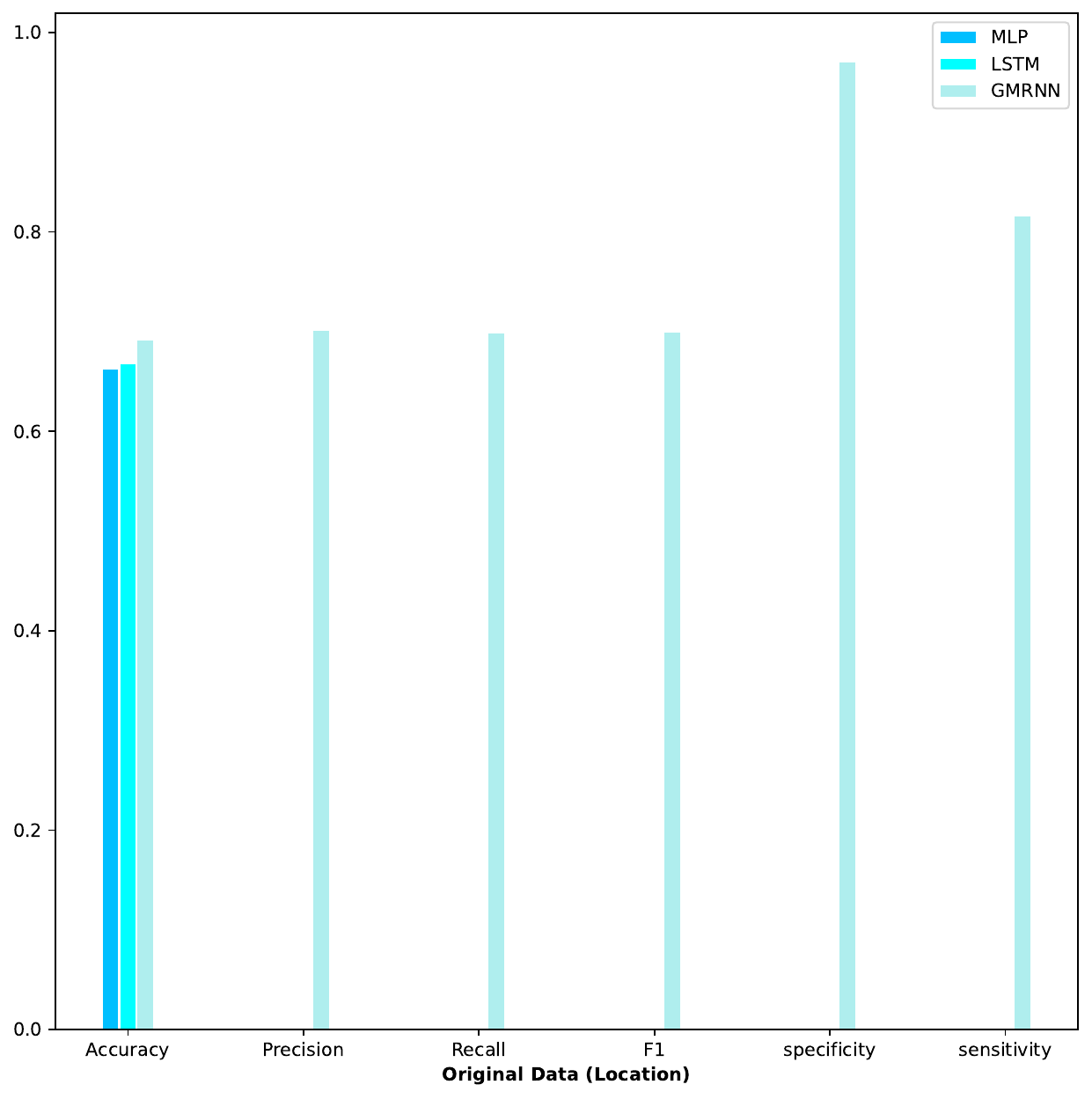}}
\subfloat[Original Data Image]{\includegraphics[width=5cm,height=5cm,]{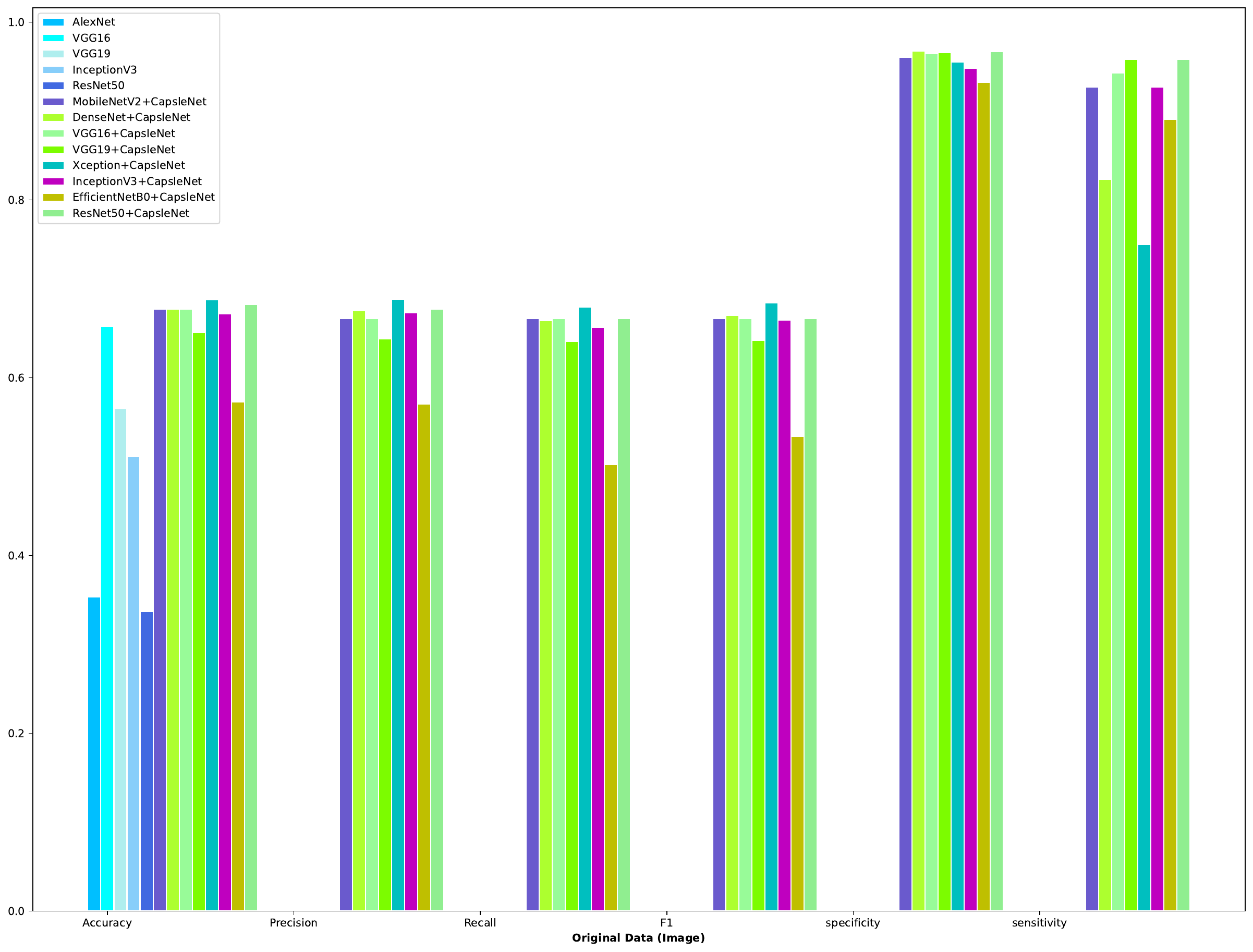}}
\subfloat[Original Data Image and Location]{\includegraphics[width=5cm,height=5cm,]{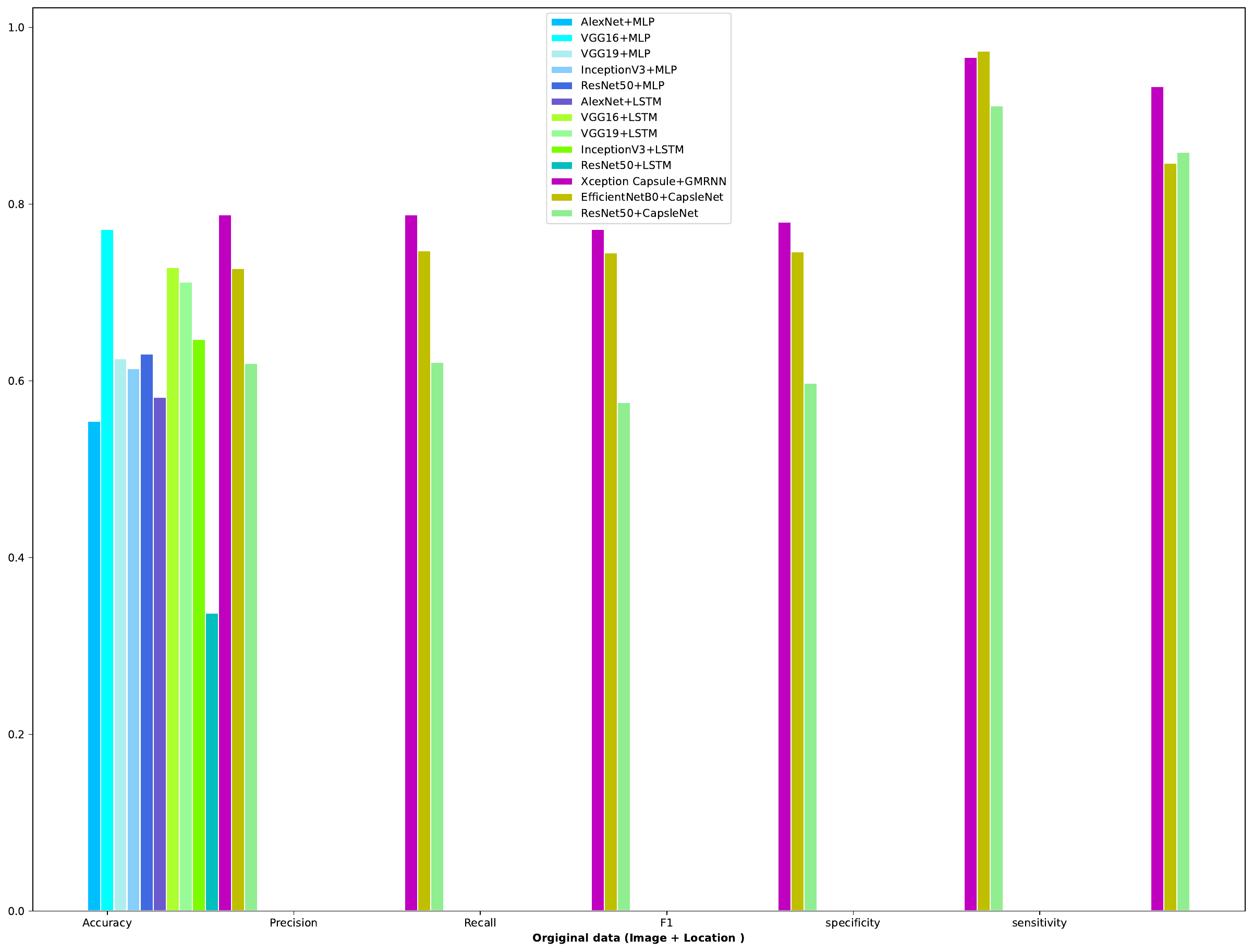}}\\
\subfloat[Augmented Data Location]{\includegraphics[width=5cm,height=5cm,]{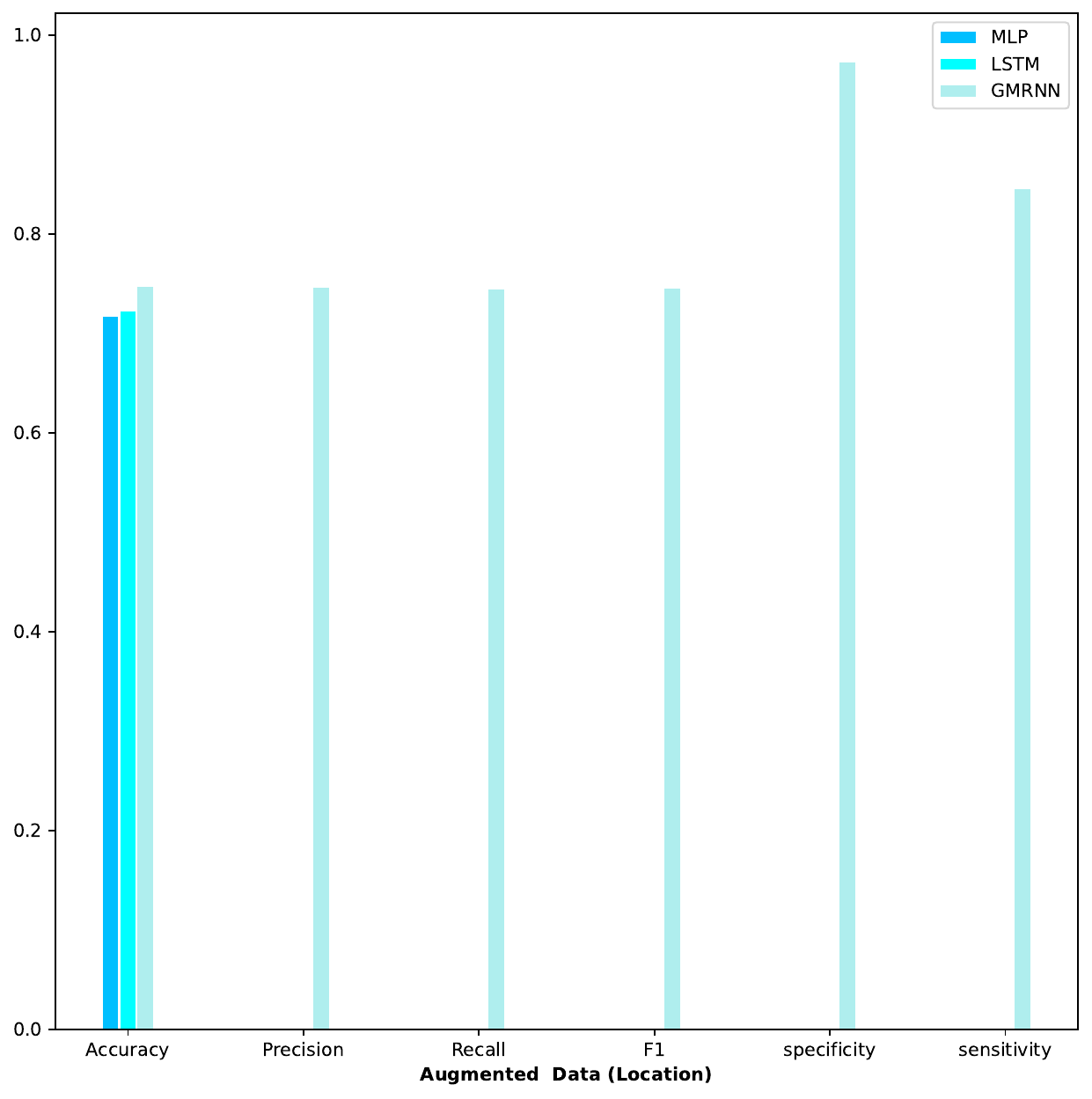}}
\subfloat[Augmented Data Image]{\includegraphics[width=5cm,height=5cm,]{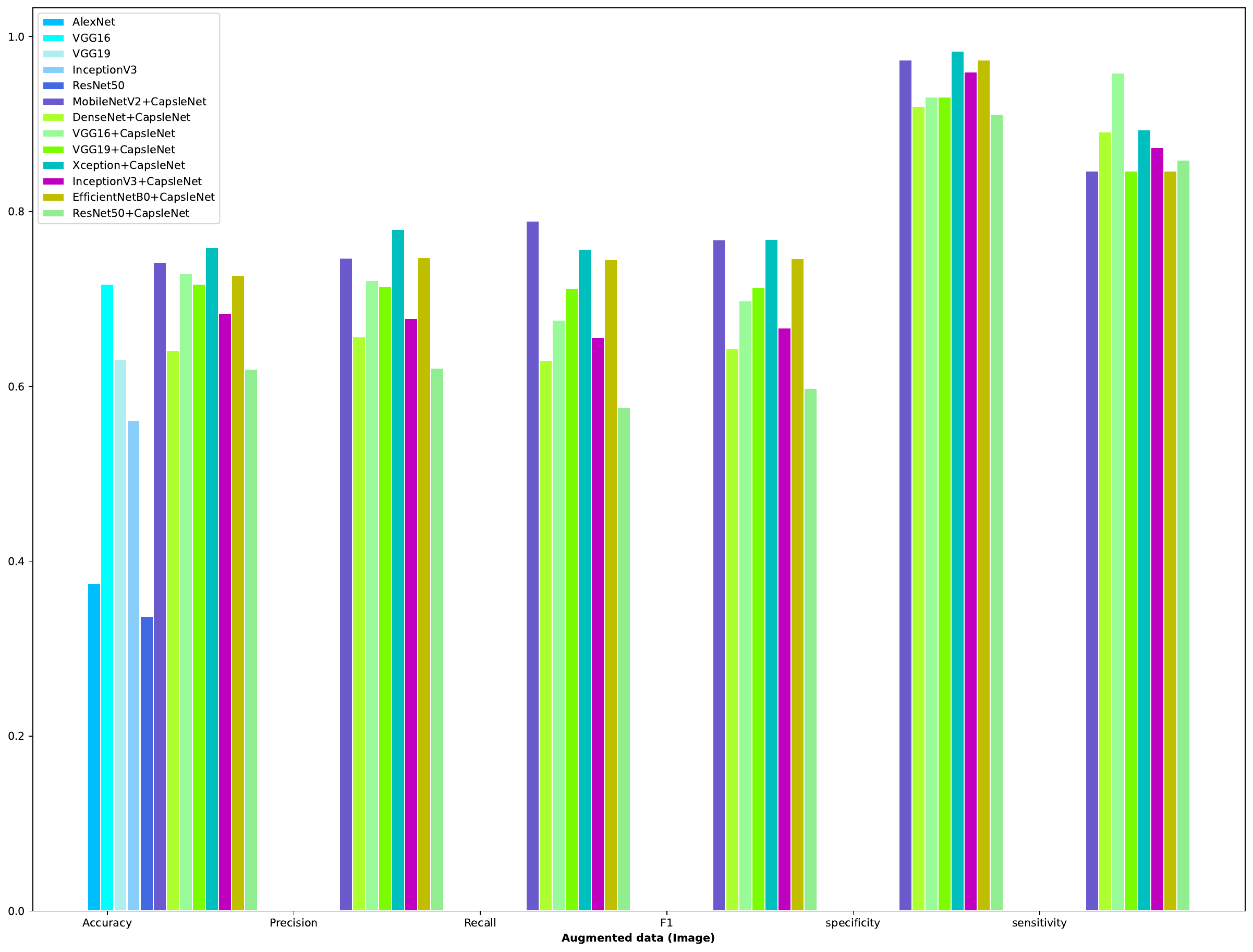}}
\subfloat[Augmented Data Image and Location]{\includegraphics[width=5cm,height=5cm,]{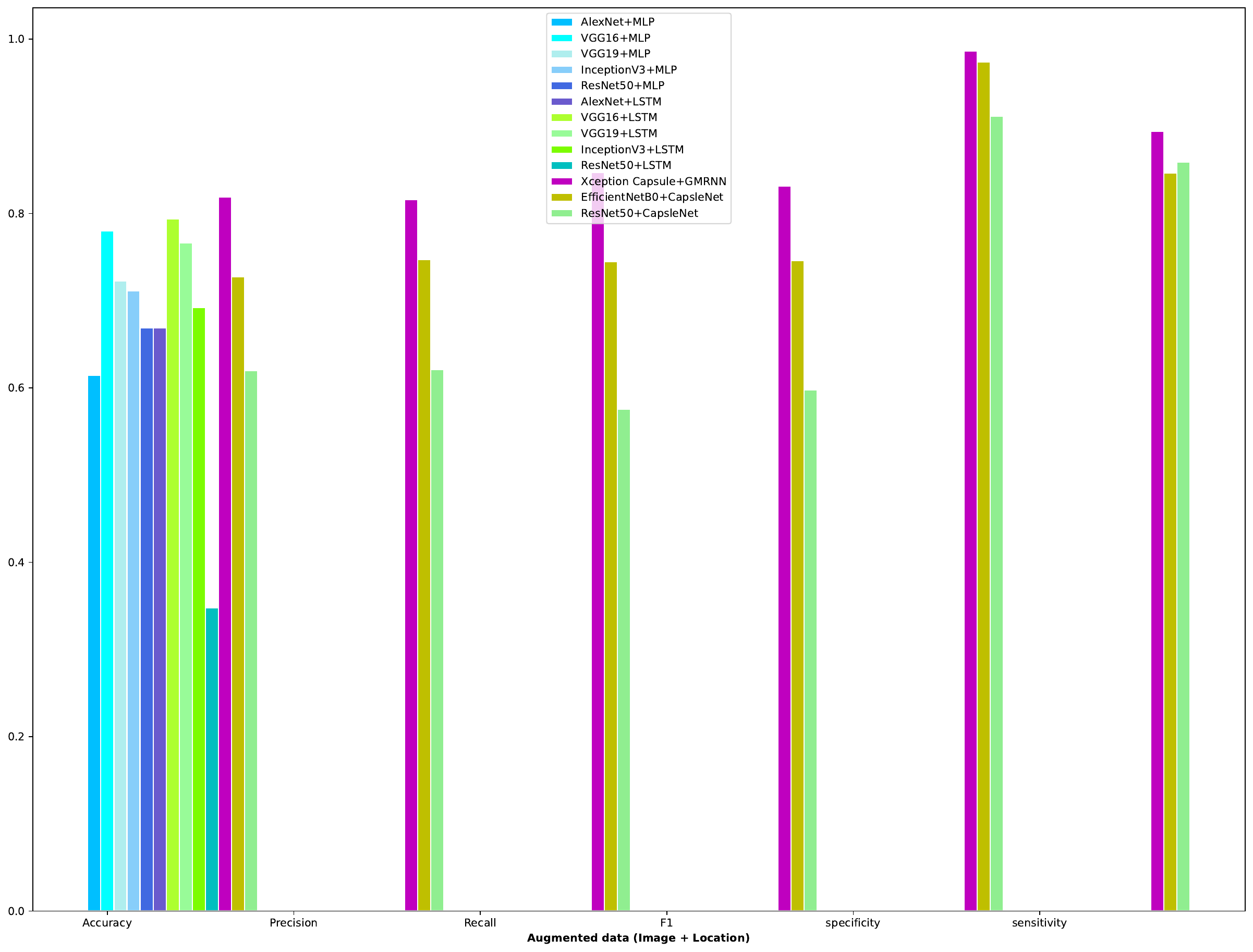}}

\caption{Bar plot for four wound class classification (D vs. P vs. S vs. V) on AZH dataset with original body map.}
\label{Fig5}
\end{figure*}

\begin{figure*}
\centering
\subfloat[Original Data Location]{\includegraphics[width=5cm,height=5cm]{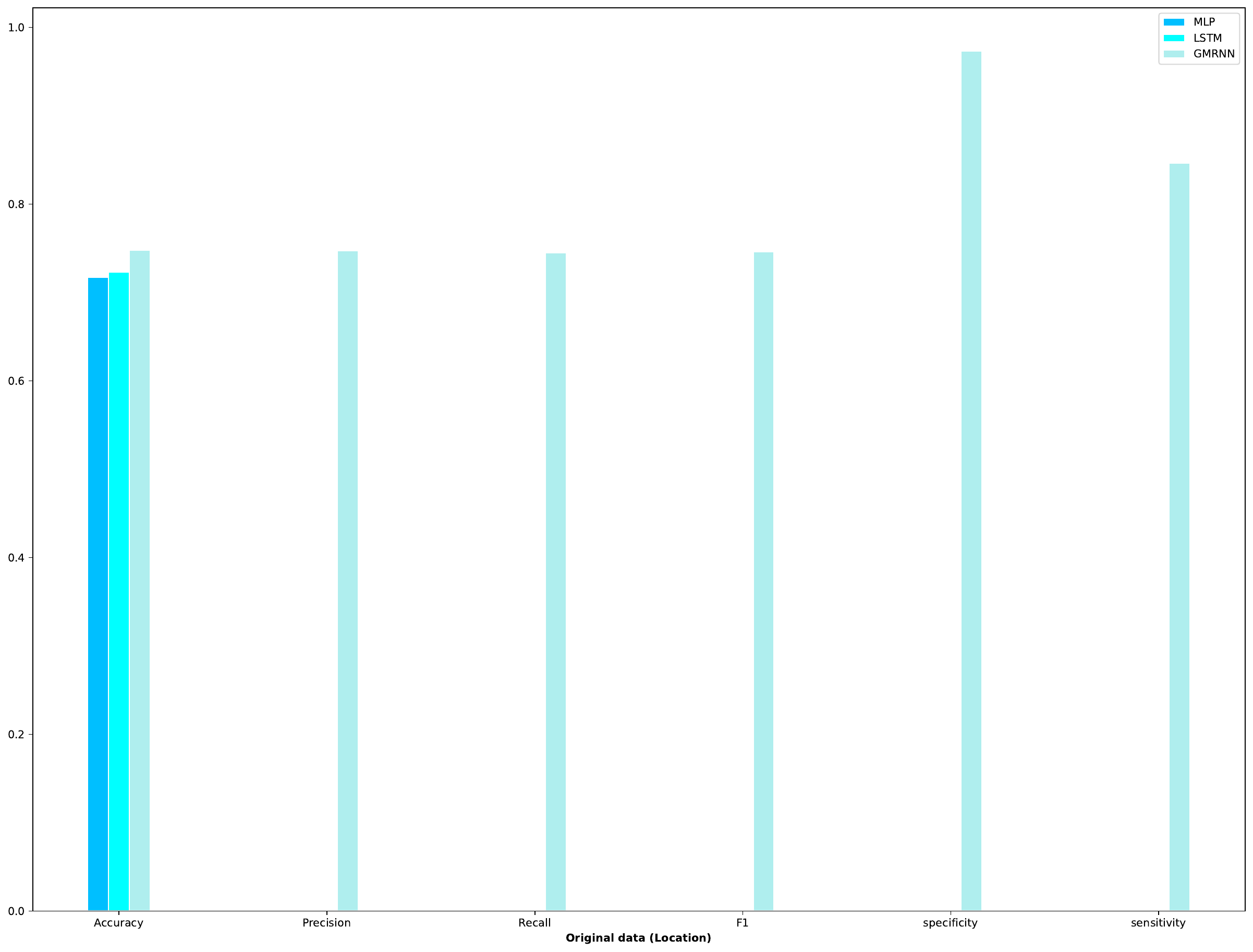}}
\subfloat[Orginal Data Image and Location]{\includegraphics[width=5cm,height=5cm,]{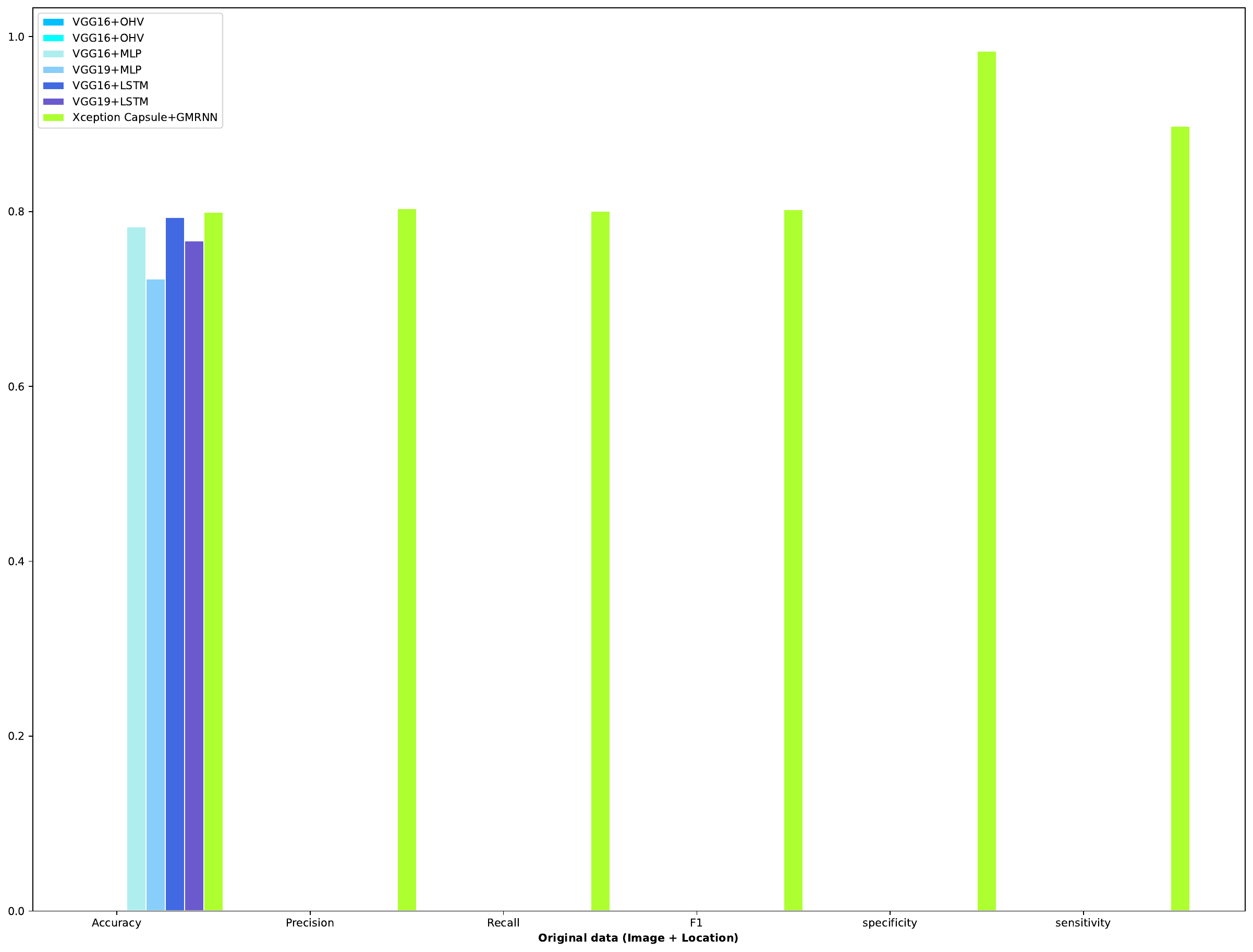}}
\subfloat[Augmented Data Location]{\includegraphics[width=5cm,height=5cm,]{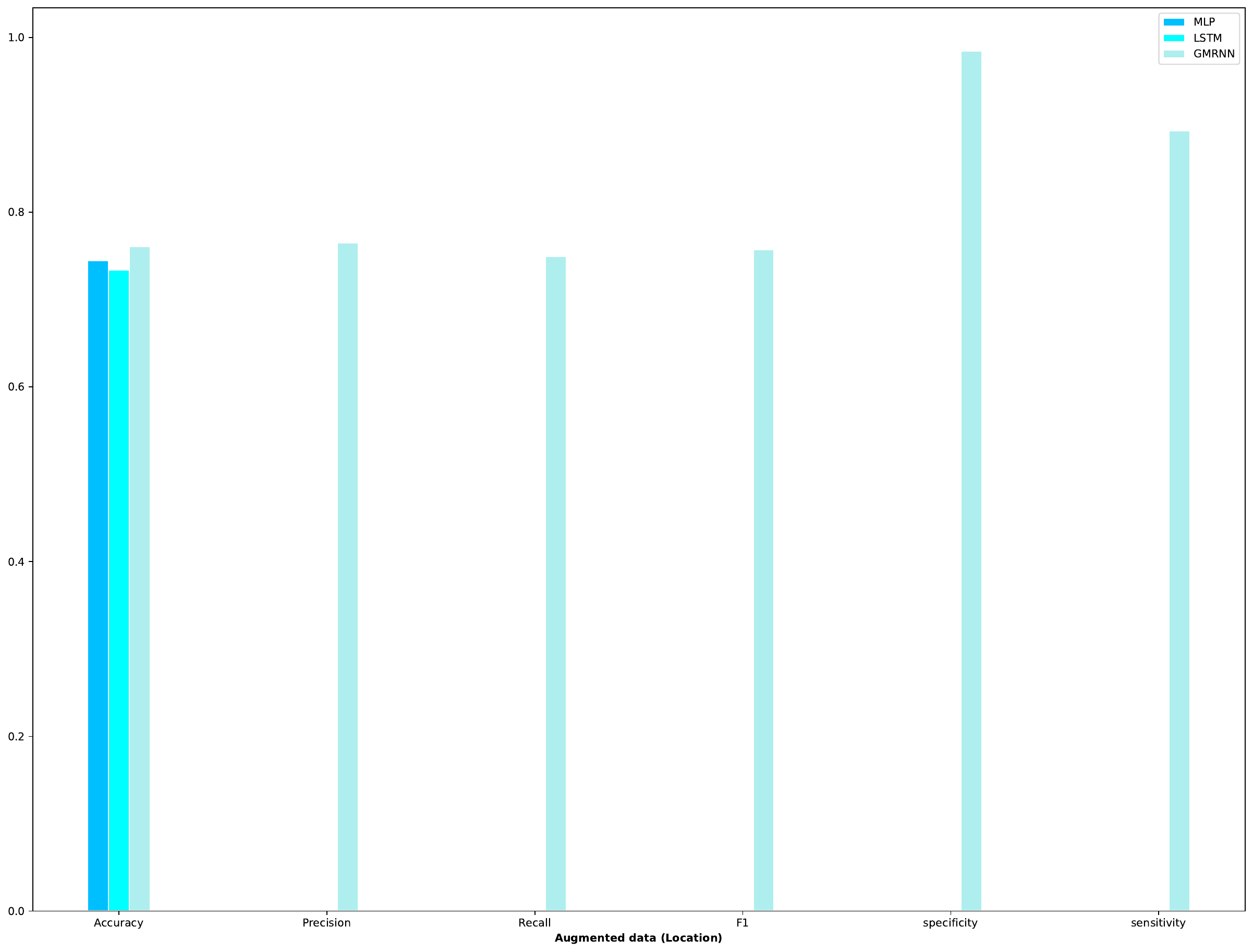}}\\
\subfloat[Augmented Data Image and Location]{\includegraphics[width=5cm,height=5cm,]{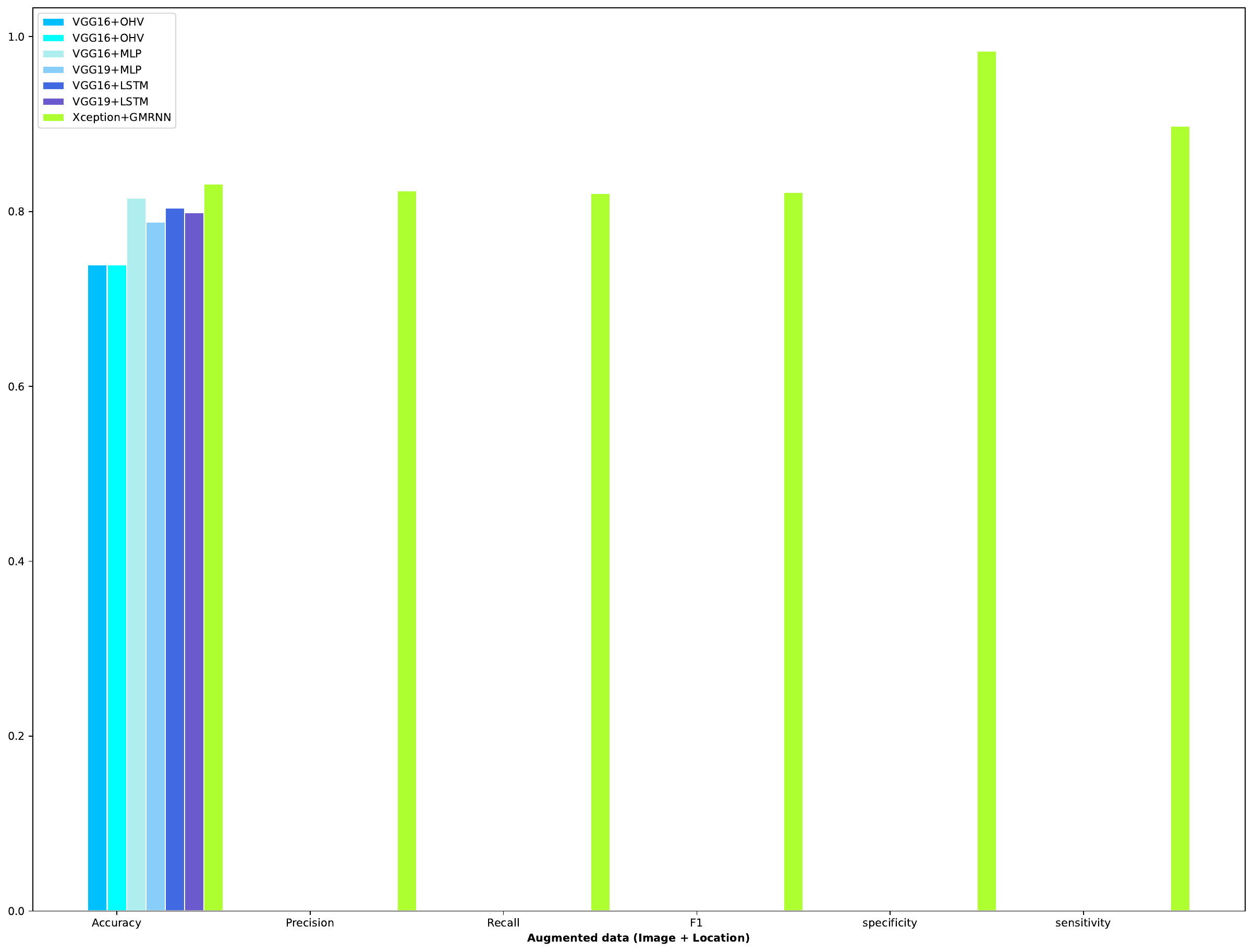}}
\caption{Bar plot for wound class classification (D vs. P vs. S vs. V) on AZH dataset with simplified body map.}
\label{Fig6}
\end{figure*}

The same four wound-class classification (D vs. P vs. S vs. V) on the AZH dataset was performed using the simplified body map, which includes 323 locations. Table \ref{Table2} presents the results of these experiment results on the AZH dataset with the simplified body map. Since the proposed framework is unaffected by changes in the body map, it was excluded from Table \ref{Table2}. With improved accuracy across all models, we used the simplified body map for all subsequent experiments. To further analyze, bar plots for four wound-class classification (D vs. P vs. S vs. V) on AZH dataset with an original body map and a simplified body map, are presented in Figure \ref{Fig5} and Figure \ref{Fig6}.

\begin{table*}[]

\centering
\resizebox{\textwidth}{!}{
\begin{tabular}{|l|l|llllll|lllll|l|}
\hline
                                  &                 & \multicolumn{1}{l|}{Accuracy} & \multicolumn{1}{l|}{Precision} & \multicolumn{1}{l|}{Recall} & \multicolumn{1}{l|}{F1}     & \multicolumn{1}{l|}{specificity} & sensitivity & \multicolumn{1}{l|}{Accuracy} & \multicolumn{1}{l|}{Precision} & \multicolumn{1}{l|}{Recall} & \multicolumn{1}{l|}{F1}     & specificity & sensitivity \\ \hline
\multirow{3}{*}{Location}         &                 & \multicolumn{6}{c|}{Original Data}                                                                                                                                          & \multicolumn{5}{c|}{Augmented data}                                                                                                      &             \\ \cline{2-14}
                                  & MLP             & \multicolumn{1}{l|}{0.7174}   & \multicolumn{1}{l|}{-}         & \multicolumn{1}{l|}{-}      & \multicolumn{1}{l|}{-}      & \multicolumn{1}{l|}{-}           & -           & \multicolumn{1}{l|}{0.7446}   & \multicolumn{1}{l|}{-}         & \multicolumn{1}{l|}{-}      & \multicolumn{1}{l|}{-}      & -           & -           \\ \cline{2-14}
                                  & LSTM            & \multicolumn{1}{l|}{0.7228}   & \multicolumn{1}{l|}{-}         & \multicolumn{1}{l|}{-}      & \multicolumn{1}{l|}{-}      & \multicolumn{1}{l|}{-}           & -           & \multicolumn{1}{l|}{0.7337}   & \multicolumn{1}{l|}{-}         & \multicolumn{1}{l|}{-}      & \multicolumn{1}{l|}{-}      & -           & -           \\ \cline{2-14}
                                  & GMRNN           & \multicolumn{1}{l|}{0.7479}   & \multicolumn{1}{l|}{0.7473}    & \multicolumn{1}{l|}{0.7449} & \multicolumn{1}{l|}{0.7461} & \multicolumn{1}{l|}{0.9735}      & 0.8462      & \multicolumn{1}{l|}{0.7607}   & \multicolumn{1}{l|}{0.7650}    & \multicolumn{1}{l|}{0.7571} & \multicolumn{1}{l|}{0.7571} & 0.9846      & 0.8932      \\ \hline
\multirow{7}{*}{Image + Location} & VGG16 + OHV     & \multicolumn{1}{l|}{N/A}      & \multicolumn{1}{l|}{-}         & \multicolumn{1}{l|}{-}      & \multicolumn{1}{l|}{-}      & \multicolumn{1}{l|}{-}           & -           & \multicolumn{1}{l|}{0.7727}   & \multicolumn{1}{l|}{-}         & \multicolumn{1}{l|}{-}      & \multicolumn{1}{l|}{-}      & -           & -           \\ \cline{2-14}
                                  & VGG16 + OHV     & \multicolumn{1}{l|}{N/A}      & \multicolumn{1}{l|}{-}         & \multicolumn{1}{l|}{-}      & \multicolumn{1}{l|}{-}      & \multicolumn{1}{l|}{-}           & -           & \multicolumn{1}{l|}{0.7391}   & \multicolumn{1}{l|}{-}         & \multicolumn{1}{l|}{-}      & \multicolumn{1}{l|}{-}      & -           & -           \\ \cline{2-14}
                                  & VGG16 + MLP     & \multicolumn{1}{l|}{0.7826}   & \multicolumn{1}{l|}{-}         & \multicolumn{1}{l|}{-}      & \multicolumn{1}{l|}{-}      & \multicolumn{1}{l|}{-}           & -           & \multicolumn{1}{l|}{0.8152}   & \multicolumn{1}{l|}{-}         & \multicolumn{1}{l|}{-}      & \multicolumn{1}{l|}{-}      & -           & -           \\ \cline{2-14}
                                  & VGG19 + MLP     & \multicolumn{1}{l|}{0.7228}   & \multicolumn{1}{l|}{-}         & \multicolumn{1}{l|}{-}      & \multicolumn{1}{l|}{-}      & \multicolumn{1}{l|}{-}           & -           & \multicolumn{1}{l|}{0.7880}   & \multicolumn{1}{l|}{-}         & \multicolumn{1}{l|}{-}      & \multicolumn{1}{l|}{-}      & -           & -           \\ \cline{2-14}
                                  & VGG16 +   LSTM  & \multicolumn{1}{l|}{0.7935}   & \multicolumn{1}{l|}{-}         & \multicolumn{1}{l|}{-}      & \multicolumn{1}{l|}{-}      & \multicolumn{1}{l|}{-}           & -           & \multicolumn{1}{l|}{0.8043}   & \multicolumn{1}{l|}{-}         & \multicolumn{1}{l|}{-}      & \multicolumn{1}{l|}{-}      & -           & -           \\ \cline{2-14}
                                  & VGG19 + LSTM    & \multicolumn{1}{l|}{0.7663}   & \multicolumn{1}{l|}{-}         & \multicolumn{1}{l|}{-}      & \multicolumn{1}{l|}{-}      & \multicolumn{1}{l|}{-}           & -           & \multicolumn{1}{l|}{0.7989}   & \multicolumn{1}{l|}{-}         & \multicolumn{1}{l|}{-}      & \multicolumn{1}{l|}{-}      & -           & -           \\ \cline{2-14}
                                  & Xception +GMRNN & \multicolumn{1}{l|}{0.7991}   & \multicolumn{1}{l|}{0.8037}    & \multicolumn{1}{l|}{0.8005} & \multicolumn{1}{l|}{0.8020} & \multicolumn{1}{l|}{0.9838}      & 0.8974      & \multicolumn{1}{l|}{0.8312}   & \multicolumn{1}{l|}{0.8237}    & \multicolumn{1}{l|}{0.8220} & \multicolumn{1}{l|}{0.8220} & 0.9838      & 0.8974      \\ \hline
\end{tabular}
}
\caption{Four wound class classification (D vs. P vs. S vs. V) on AZH dataset with simplified body map. The
bold represents the highest results/accuracy achieved for each experiment.}
\label{Table2}
\end{table*}

With a simplified body map, he performances of MLP and LSTM exhibited the similar pattern on the location data, whereas GMRNN was the best with an accuracy of 0.7479. We also conducted experiments to examine the effect of inputting the one-hot vector (OHV) into the dense layer of the CNN. The results showed values of 0.7727 for VGG16 + OHV and 0.7391 for VGG19 + OHV, highlighting the poor performance of OHV compared to MLP and LSTM. Once again, the Xception+ GMRNN combination outperformed others in the four wound-class classification with the simplified body map.
The combination of VGG16 + MLP and VGG16 + LSTM showed an accuracy of 0.7826 and 0.7935 on image and location modalities, respectively. The Xception+ GMRNN performance showed values of 0.8189, 0.8159, 0.8469, and 0.8311, for the metrics of accuracy, precision, recall, and F1 respectively, on augmented data (as shown in Table \ref{Table1}). Similar results were observed with simplified body map on augmented data for Xception+ GMRNN.

Another experiment was conducted for wound classification among all classes in the AZH dataset. Table \ref{Table3} presents the results of this six-class classification (BG vs. N vs. D vs. P vs. S vs. V). Our proposed multi-modal approach achieved the highest accuracy of 83\% using the Xception + GMRNN combination. In comparison, the other combinations—VGG16 + MLP, VGG19 + MLP, and VGG16 + LSTM—achieved accuracies of 79.49\%, 82.48\%, and 79.49\%, respectively.

\begin{table*}[]

\centering
\begin{tabular}{|l|l|l|}
\hline
Input                            & Model           & Accuracy \\ \hline
\multirow{3}{*}{Location}        & MLP             & 0.6496   \\ \cline{2-3}
                                 & LSTM            & 0.6752   \\ \cline{2-3}
                                 & GMRNN           & 0.712    \\ \hline
\multirow{3}{*}{Image}           & VGG16           & 0.7564   \\ \cline{2-3}
                                 & VGG19           & 0.6496   \\ \cline{2-3}
                                 & Xception        & 0.779    \\ \hline
\multirow{5}{*}{Image+ location} & VGG16+MLP       & 0.7949   \\ \cline{2-3}
                                 & VGG19+MLP       & 0.8248   \\ \cline{2-3}
                                 & VGG16+LSTM      & 0.7949   \\ \cline{2-3}
                                 & VGG19+LSTM      & 0.7222   \\ \cline{2-3}
                                 & Xception+ GMRNN & 0.83     \\ \hline
\end{tabular}
\caption{Six-class classification (BG vs. N vs. D vs. P vs. S vs. V) on AZH dataset. The bold represents the
highest results/accuracy achieved for each experiment.}
\label{Table3}
\end{table*}

\begin{figure*}
\centering
\subfloat[ Location]{\includegraphics[width=5cm,height=5cm]{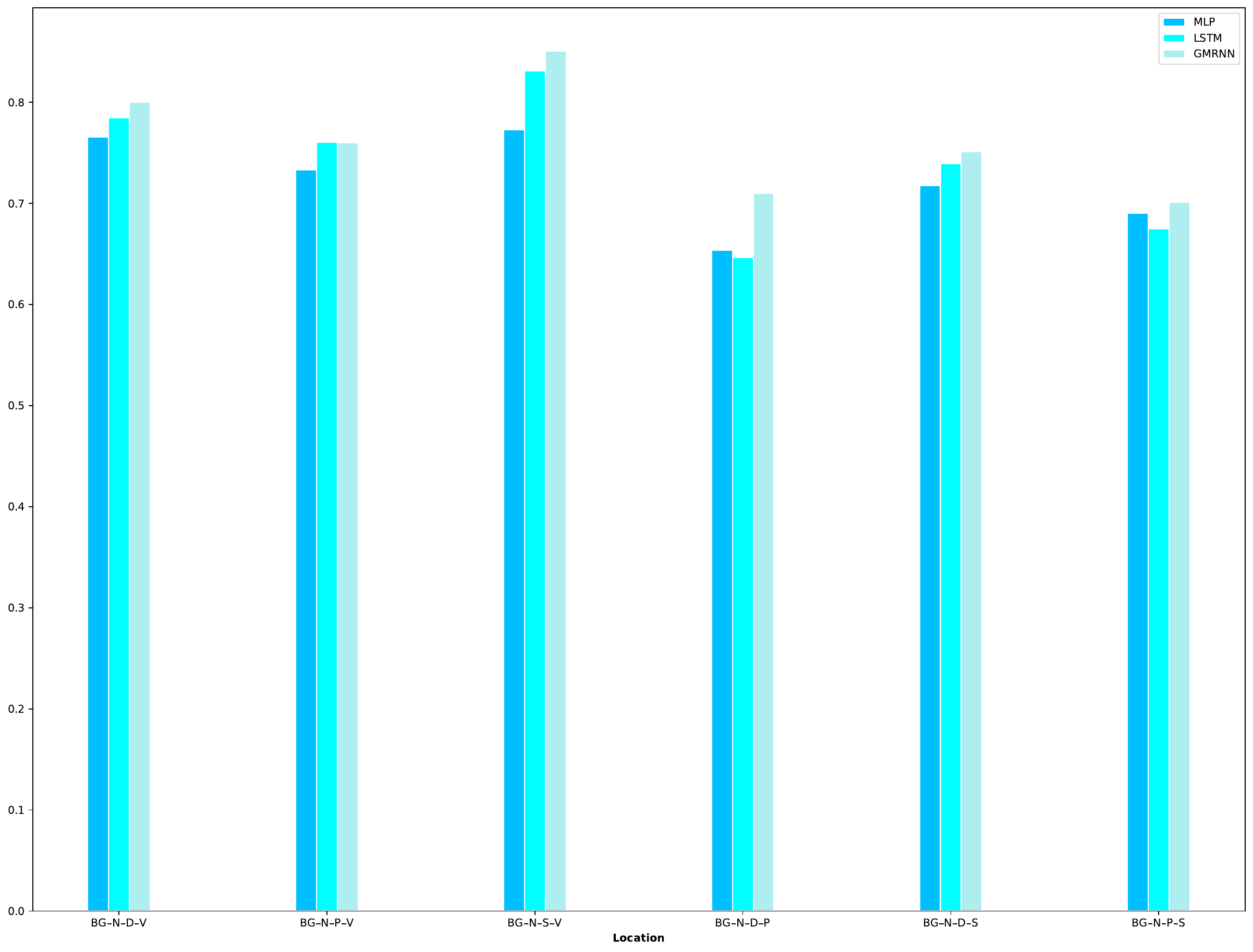}}
\subfloat[Image]{\includegraphics[width=5cm,height=5cm,]{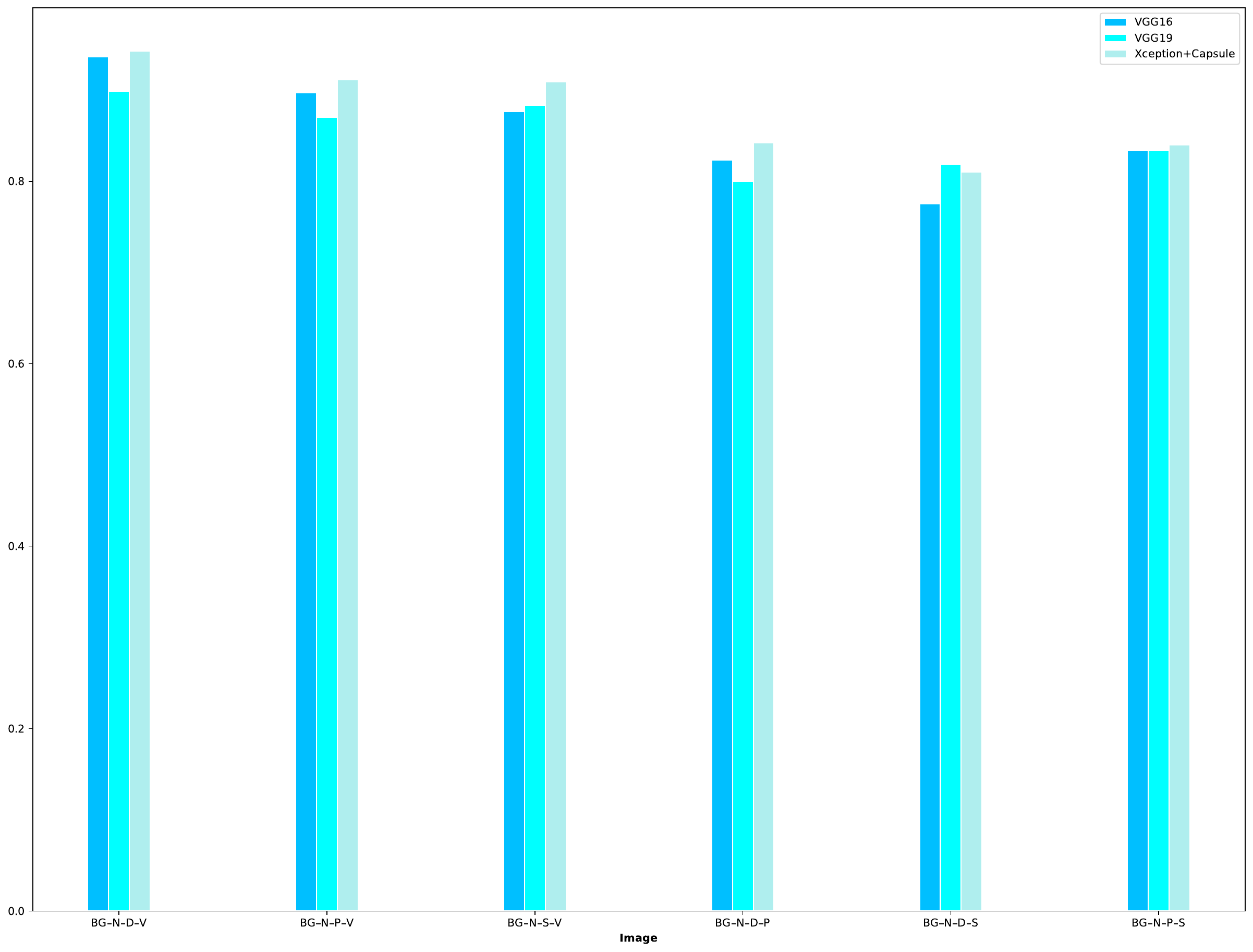}}
\subfloat[Image and Location]{\includegraphics[width=5cm,height=5cm,]{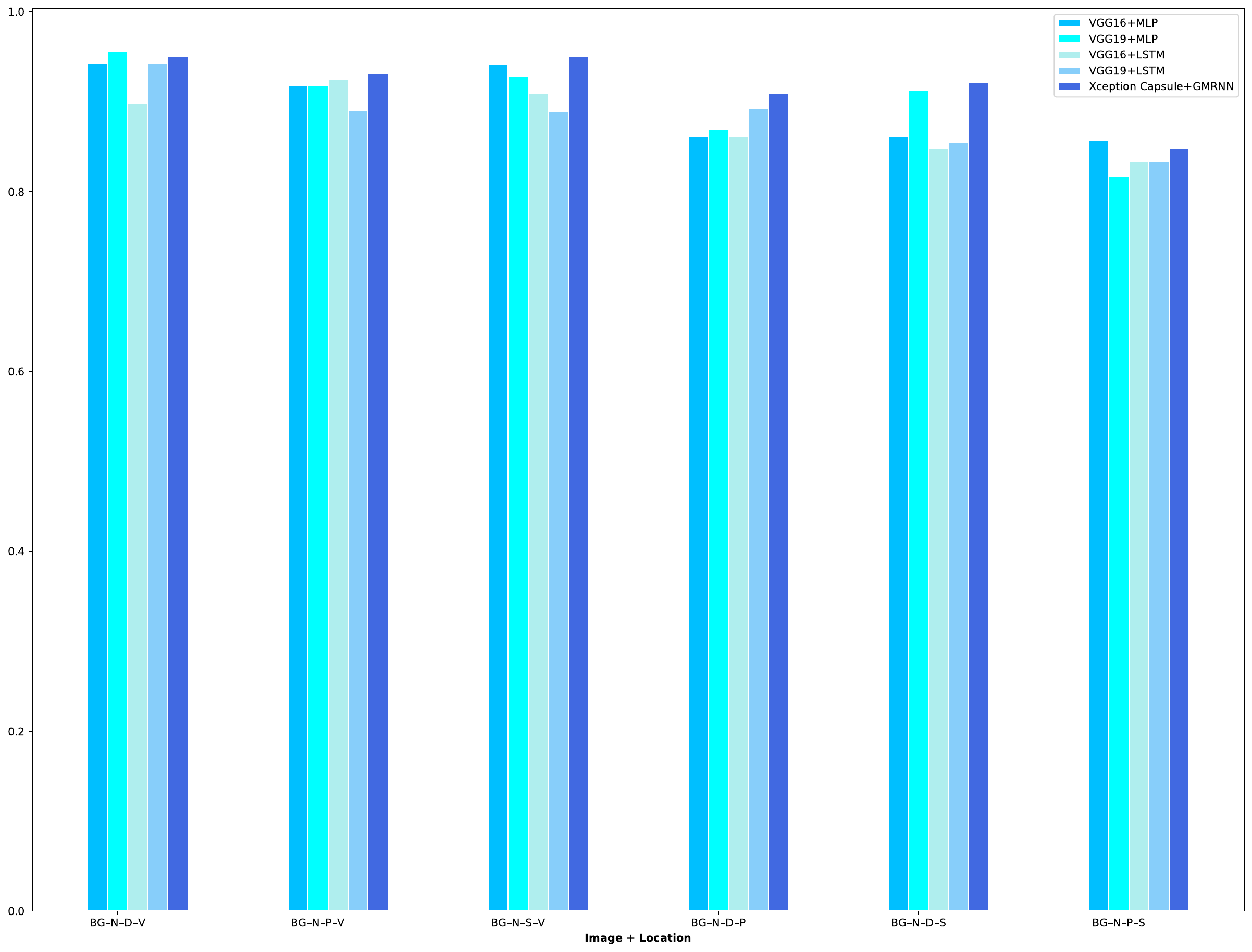}}\\
\caption{Bar plot for four five-class classifications on AZH dataset.}
\label{Imag7}
\end{figure*}

We conducted four five-class classifications on the AZH dataset. These classifications included: (1) BG vs. N vs. D vs. P vs. V, (2) BG vs. N vs. D vs. S vs. V, (3) BG vs. N vs. D vs. P vs. S, and (4) BG vs. N vs. P vs. S vs. V. Detailed results of these classifications are provided in Table \ref{Table4}. The highest accuracies were achieved using the Xception+GMRNN combination, with scores of 0.8885, 0.9310, 0.8712, and 0.8712 for classifications (1), (2), (3), and (4) respectively. The multi-modal framework consistently achieved the highest accuracy across all four classifications. Figure \ref{Imag7} illustrates the bar plots for four five-class classifications on AZH dataset for further analysis.

\begin{table*}[]

\centering
\resizebox{\textwidth}{!}{
\begin{tabular}{|l|l|llll|}
\hline
Input                            & Model           & \multicolumn{1}{l|}{BG–N–D–P–V} & \multicolumn{1}{l|}{BG–N–D–S–V} & \multicolumn{1}{l|}{BG–N–D–P–S} & BG–N–P–S–V \\ \hline
                                 &                 & \multicolumn{4}{c|}{Accuracy}                                                                                    \\ \hline
\multirow{3}{*}{Location}        & MLP             & \multicolumn{1}{l|}{0.6771}     & \multicolumn{1}{l|}{0.7500}     & \multicolumn{1}{l|}{0.5930}     & 0.6968     \\ \cline{2-6}
                                 & LSTM            & \multicolumn{1}{l|}{0.6875}     & \multicolumn{1}{l|}{0.7200}     & \multicolumn{1}{l|}{0.5930}     & 0.7181     \\ \cline{2-6}
                                 & GMRNN           & \multicolumn{1}{l|}{0.6920}     & \multicolumn{1}{l|}{0.7420}     & \multicolumn{1}{l|}{0.6230}     & 0.7050     \\ \hline
\multirow{3}{*}{Image}           & VGG16           & \multicolumn{1}{l|}{0.6979}     & \multicolumn{1}{l|}{0.7050}     & \multicolumn{1}{l|}{0.6453}     & 0.7553     \\ \cline{2-6}
                                 & VGG19           & \multicolumn{1}{l|}{0.7656}     & \multicolumn{1}{l|}{0.7450}     & \multicolumn{1}{l|}{0.6744}     & 0.7234     \\ \cline{2-6}
                                 & Xception        & \multicolumn{1}{l|}{0.7774}     & \multicolumn{1}{l|}{0.7723}     & \multicolumn{1}{l|}{0.7701}     & 0.7610     \\ \hline
\multirow{5}{*}{Image+ location} & VGG16+MLP       & \multicolumn{1}{l|}{0.8646}     & \multicolumn{1}{l|}{0.8500}     & \multicolumn{1}{l|}{0.8314}     & 0.8404     \\ \cline{2-6}
                                 & VGG19+MLP       & \multicolumn{1}{l|}{0.8542}     & \multicolumn{1}{l|}{0.8650}     & \multicolumn{1}{l|}{0.7733}     & 0.8617     \\ \cline{2-6}
                                 & VGG16+LSTM      & \multicolumn{1}{l|}{0.8438}     & \multicolumn{1}{l|}{0.9100}     & \multicolumn{1}{l|}{0.7733}     & 0.7713     \\ \cline{2-6}
                                 & VGG19+LSTM      & \multicolumn{1}{l|}{0.8438}     & \multicolumn{1}{l|}{0.9100}     & \multicolumn{1}{l|}{0.7733}     & 0.7713     \\ \cline{2-6}
                                 & Xception+ GMRNN & \multicolumn{1}{l|}{0.8885}     & \multicolumn{1}{l|}{0.9310}     & \multicolumn{1}{l|}{0.8712}     & 0.8712     \\ \hline
\end{tabular}
}
\caption{Four five-class classifications on AZH dataset. The bold represents the highest results/accuracy
achieved for each experiment.}
\label{Table4}
\end{table*}

We also performed six four-class classifications and one wound class classification on the AZH dataset, as detailed in Tables \ref{Table5} and \ref{Table6}. The classifications were: (1) BG vs. N vs. D vs. V, (2) BG vs. N vs. P vs. V, (3) BG vs. N vs. S vs. V, (4) BG vs. N vs. D vs. P, (5) BG vs. N vs. D vs. S, and (6) BG vs. N vs. P vs. S. The highest accuracies achieved were 95.09\%, 93.12\%, 95.01\%, 90.99\%, 92.12\%, and 84.84\% for classifications (1), (2), (3), (4), (5), and (6), respectively. The proposed multi-modal framework again achieved the highest accuracy across all six classifications. Detailed results are shown in Table \ref{Table5}.

\begin{table*}[]
\centering
\resizebox{\textwidth}{!}{
\begin{tabular}{|l|l|llllll|}
\hline
Input                            & Model           & \multicolumn{1}{l|}{BG–N–D–V} & \multicolumn{1}{l|}{BG–N–P–V} & \multicolumn{1}{l|}{BG–N–S–V} & \multicolumn{1}{l|}{BG–N–D–P} & \multicolumn{1}{l|}{BG–N–D–S} & BG–N–P–S \\ \hline
                                 &                 & \multicolumn{6}{c|}{Accuracy}                                                                                                                                            \\ \hline
\multirow{3}{*}{Location}        & MLP             & \multicolumn{1}{l|}{0.7658}   & \multicolumn{1}{l|}{0.7329}   & \multicolumn{1}{l|}{0.7727}   & \multicolumn{1}{l|}{0.6538}   & \multicolumn{1}{l|}{0.7174}   & 0.6904   \\ \cline{2-8}
                                 & LSTM            & \multicolumn{1}{l|}{0.7848}   & \multicolumn{1}{l|}{0.7603}   & \multicolumn{1}{l|}{0.8312}   & \multicolumn{1}{l|}{0.6462}   & \multicolumn{1}{l|}{0.7391}   & 0.6746   \\ \cline{2-8}
                                 & GMRNN           & \multicolumn{1}{l|}{0.8000}   & \multicolumn{1}{l|}{0.7601}   & \multicolumn{1}{l|}{0.8509}   & \multicolumn{1}{l|}{0.7101}   & \multicolumn{1}{l|}{0.7511}   & 0.7009   \\ \hline
\multirow{3}{*}{Image}           & VGG16           & \multicolumn{1}{l|}{0.9367}   & \multicolumn{1}{l|}{0.8973}   & \multicolumn{1}{l|}{0.8766}   & \multicolumn{1}{l|}{0.8231}   & \multicolumn{1}{l|}{0.7754}   & 0.8333   \\ \cline{2-8}
                                 & VGG19           & \multicolumn{1}{l|}{0.8987}   & \multicolumn{1}{l|}{0.8699}   & \multicolumn{1}{l|}{0.8831}   & \multicolumn{1}{l|}{0.8000}   & \multicolumn{1}{l|}{0.8188}   & 0.8333   \\ \cline{2-8}
                                 & Xception        & \multicolumn{1}{l|}{0.943}    & \multicolumn{1}{l|}{0.9111}   & \multicolumn{1}{l|}{90.91}    & \multicolumn{1}{l|}{0.8422}   & \multicolumn{1}{l|}{0.8101}   & 0.8401   \\ \hline
\multirow{5}{*}{Image+ location} & VGG16+MLP       & \multicolumn{1}{l|}{0.9430}   & \multicolumn{1}{l|}{0.9178}   & \multicolumn{1}{l|}{0.9416}   & \multicolumn{1}{l|}{0.8615}   & \multicolumn{1}{l|}{0.8615}   & 0.8571   \\ \cline{2-8}
                                 & VGG19+MLP       & \multicolumn{1}{l|}{0.9557}   & \multicolumn{1}{l|}{0.9178}   & \multicolumn{1}{l|}{0.9286}   & \multicolumn{1}{l|}{0.8692}   & \multicolumn{1}{l|}{0.9130}   & 0.8175   \\ \cline{2-8}
                                 & VGG16+LSTM      & \multicolumn{1}{l|}{0.8987}   & \multicolumn{1}{l|}{0.9247}   & \multicolumn{1}{l|}{0.9091}   & \multicolumn{1}{l|}{0.8615}   & \multicolumn{1}{l|}{0.8478}   & 0.8333   \\ \cline{2-8}
                                 & VGG19+LSTM      & \multicolumn{1}{l|}{0.9430}   & \multicolumn{1}{l|}{0.8904}   & \multicolumn{1}{l|}{0.8889}   & \multicolumn{1}{l|}{0.8923}   & \multicolumn{1}{l|}{0.8551}   & 0.8333   \\ \cline{2-8}
                                 & Xception+ GMRNN & \multicolumn{1}{l|}{0.9509}   & \multicolumn{1}{l|}{0.9312}   & \multicolumn{1}{l|}{95.01}    & \multicolumn{1}{l|}{90.99}    & \multicolumn{1}{l|}{0.9212}   & 0.8484   \\ \hline
\end{tabular}
}
\caption{Six four-class classifications on AZH dataset. The bold represents the highest results/accuracy
achieved for each experiment.}
\label{Table5}
\end{table*}

\begin{figure*}
\centering
\subfloat[ Location]{\includegraphics[width=5cm,height=5cm]{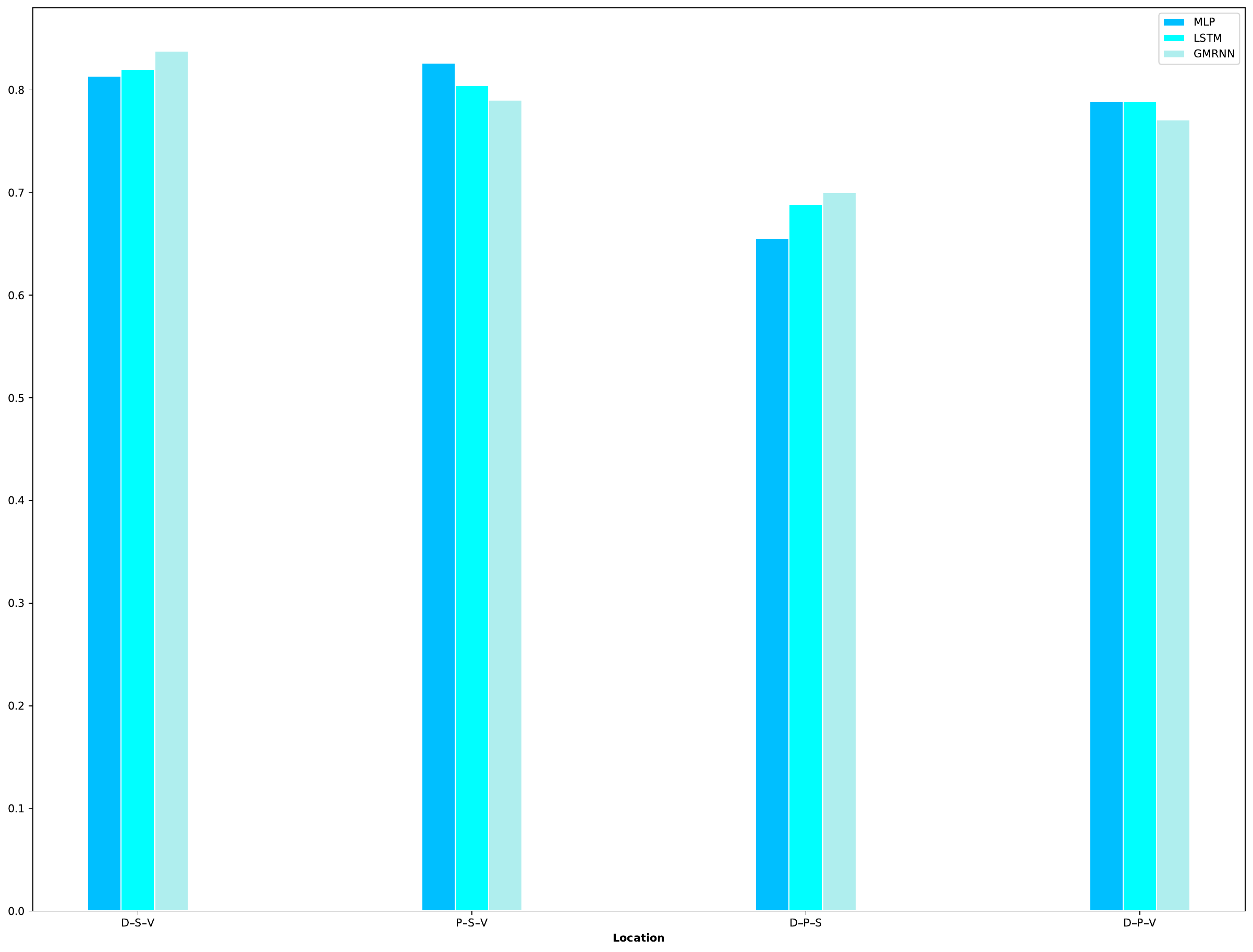}}
\subfloat[Image]{\includegraphics[width=5cm,height=5cm,]{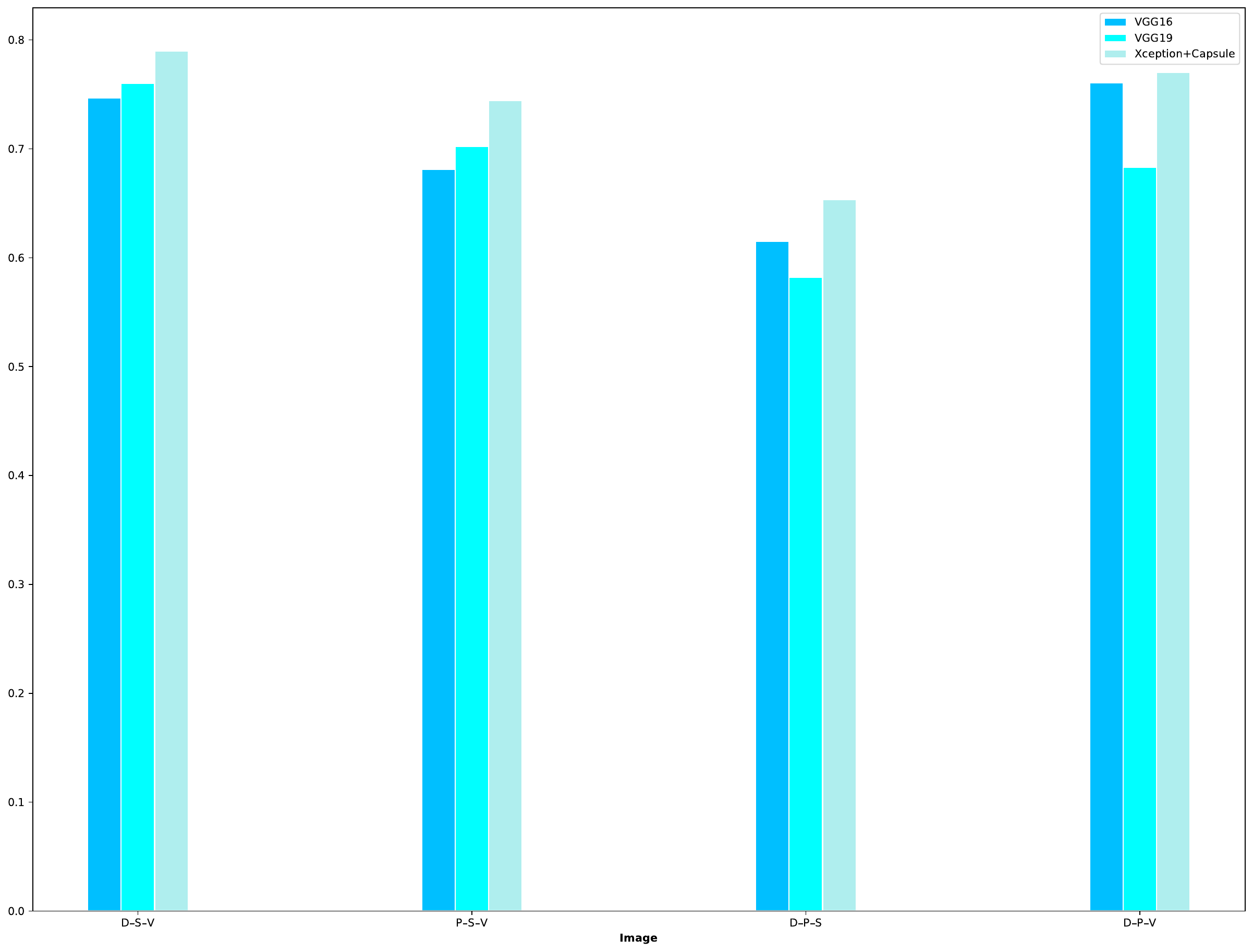}}
\subfloat[Image and Location]{\includegraphics[width=5cm,height=5cm,]{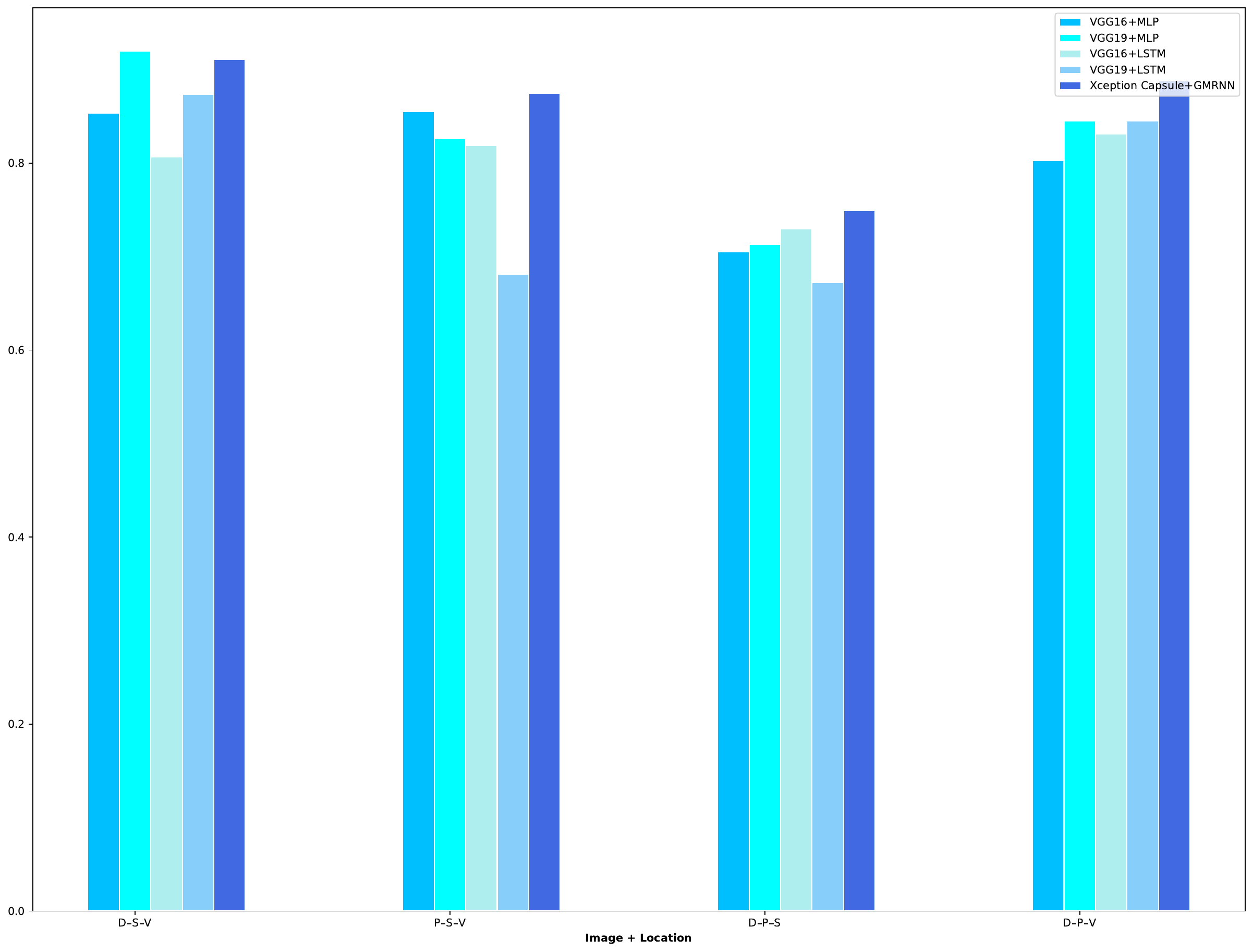}}\\
\caption{Bar plot for Six three-wound-class classifications on AZH dataset. }
\label{Fig8}
\end{figure*}

\begin{table*}[]

\centering
\begin{tabular}{|l|l|llll|}
\hline
                                 &                 & \multicolumn{1}{l|}{D–S–V} & \multicolumn{1}{l|}{P–S–V} & \multicolumn{1}{l|}{D–P–S} & D–P–V \\ \hline
Input                            & Model           & \multicolumn{4}{c|}{Accuracy}                                                                \\ \hline
\multirow{3}{*}{Location}        & MLP             & \multicolumn{1}{l|}{81.33} & \multicolumn{1}{l|}{82.61} & \multicolumn{1}{l|}{65.57} & 78.87 \\ \cline{2-6}
                                 & LSTM            & \multicolumn{1}{l|}{82.00} & \multicolumn{1}{l|}{80.43} & \multicolumn{1}{l|}{68.85} & 78.87 \\ \cline{2-6}
                                 & GMRNN           & \multicolumn{1}{l|}{83.81} & \multicolumn{1}{l|}{79.01} & \multicolumn{1}{l|}{70.01} & 77.09 \\ \hline
\multirow{3}{*}{Image}           & VGG16           & \multicolumn{1}{l|}{74.67} & \multicolumn{1}{l|}{68.12} & \multicolumn{1}{l|}{61.48} & 76.06 \\ \cline{2-6}
                                 & VGG19           & \multicolumn{1}{l|}{76.00} & \multicolumn{1}{l|}{70.23} & \multicolumn{1}{l|}{58.20} & 68.31 \\ \cline{2-6}
                                 & Xception        & \multicolumn{1}{l|}{79.00} & \multicolumn{1}{l|}{74.44} & \multicolumn{1}{l|}{65.32} & 77.00 \\ \hline
\multirow{5}{*}{Image+ location} & VGG16+MLP       & \multicolumn{1}{l|}{85.33} & \multicolumn{1}{l|}{85.51} & \multicolumn{1}{l|}{70.49} & 80.28 \\ \cline{2-6}
                                 & VGG19+MLP       & \multicolumn{1}{l|}{92.00} & \multicolumn{1}{l|}{82.61} & \multicolumn{1}{l|}{71.31} & 84.51 \\ \cline{2-6}
                                 & VGG16+LSTM      & \multicolumn{1}{l|}{80.67} & \multicolumn{1}{l|}{81.88} & \multicolumn{1}{l|}{72.95} & 83.10 \\ \cline{2-6}
                                 & VGG19+LSTM      & \multicolumn{1}{l|}{87.33} & \multicolumn{1}{l|}{68.12} & \multicolumn{1}{l|}{67.21} & 84.51 \\ \cline{2-6}
                                 & Xception+ GMRNN & \multicolumn{1}{l|}{91.08} & \multicolumn{1}{l|}{87.47} & \multicolumn{1}{l|}{74.91} & 88.81 \\ \hline
\end{tabular}
\caption{Four three-wound-class classifications on AZH dataset. The bold represents the highest results/
accuracy achieved for each experiment.}
\label{Table6}
\end{table*}
Additionally, four three-wound-class classifications were performed on the AZH dataset. These classifications were: (1) D vs. S vs. V, (2) P vs. S vs. V, (3) D vs. P vs. S, and (4) D vs. P vs. V. The highest accuracies observed were 91.08\%, 87.47\%, 74.91\%, and 88.81\% for classifications (1), (2), (3), and (4) respectively. The Xception+GMRNN combination achieved the highest accuracy in all four wound-class classifications. Detailed results are presented in Table\ref{Table6}. Figure \ref{Fig8} presents the bar plot for four three-wound-class classifications on AZH dataset for further analysis.

\begin{figure*}
\centering
\subfloat[ Location]{\includegraphics[width=5cm,height=5cm]{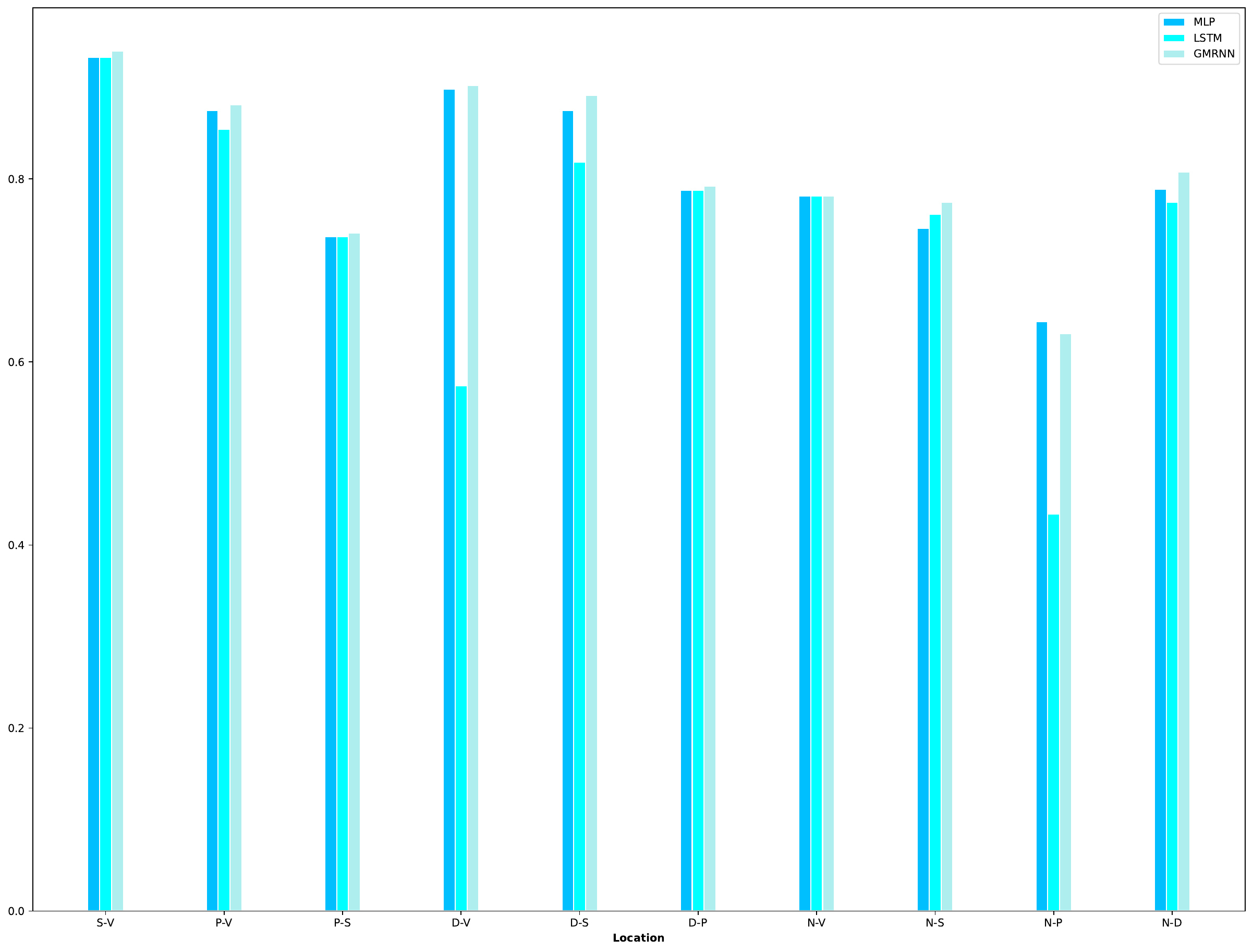}}
\subfloat[Image]{\includegraphics[width=5cm,height=5cm,]{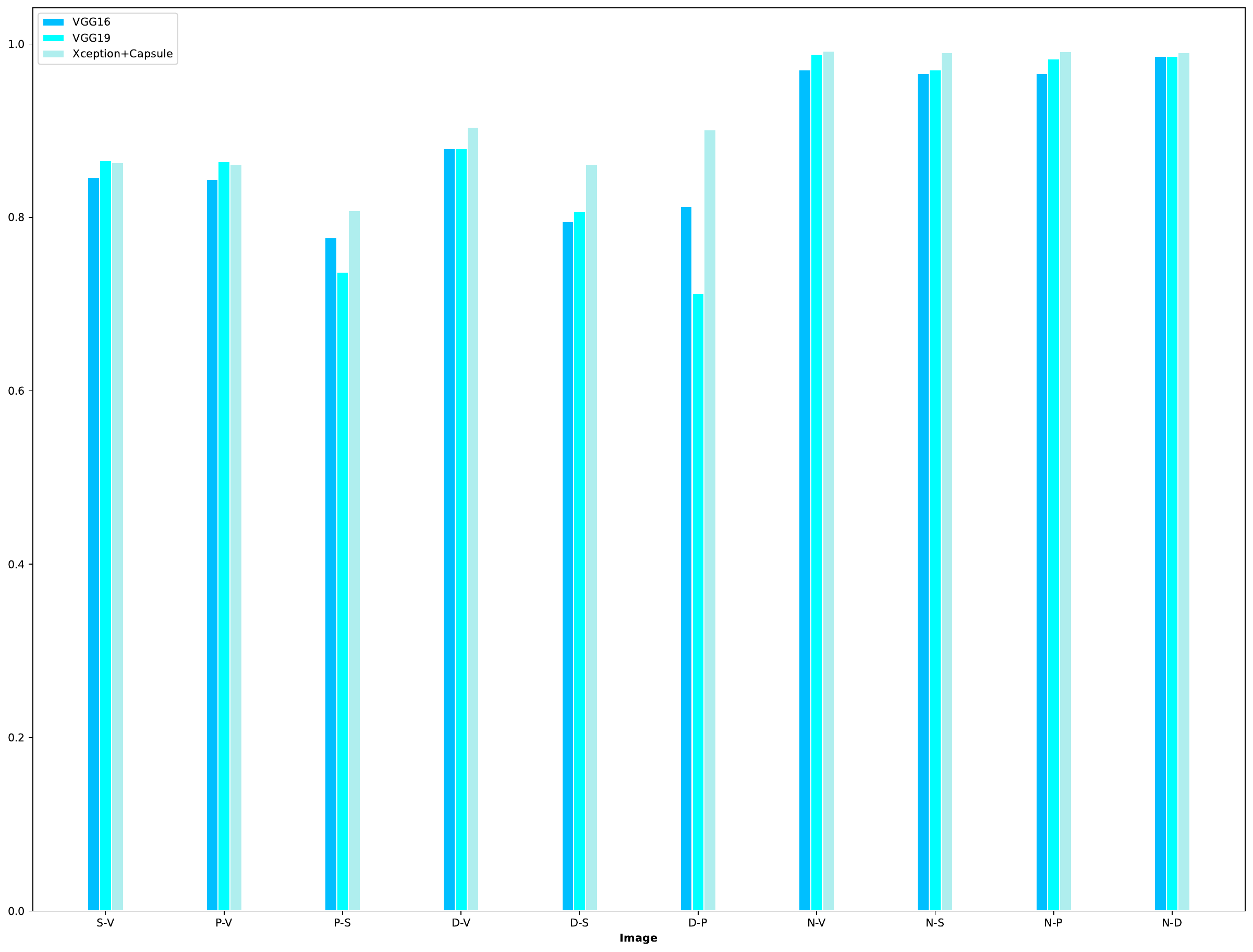}}
\subfloat[Image and Location]{\includegraphics[width=5cm,height=5cm,]{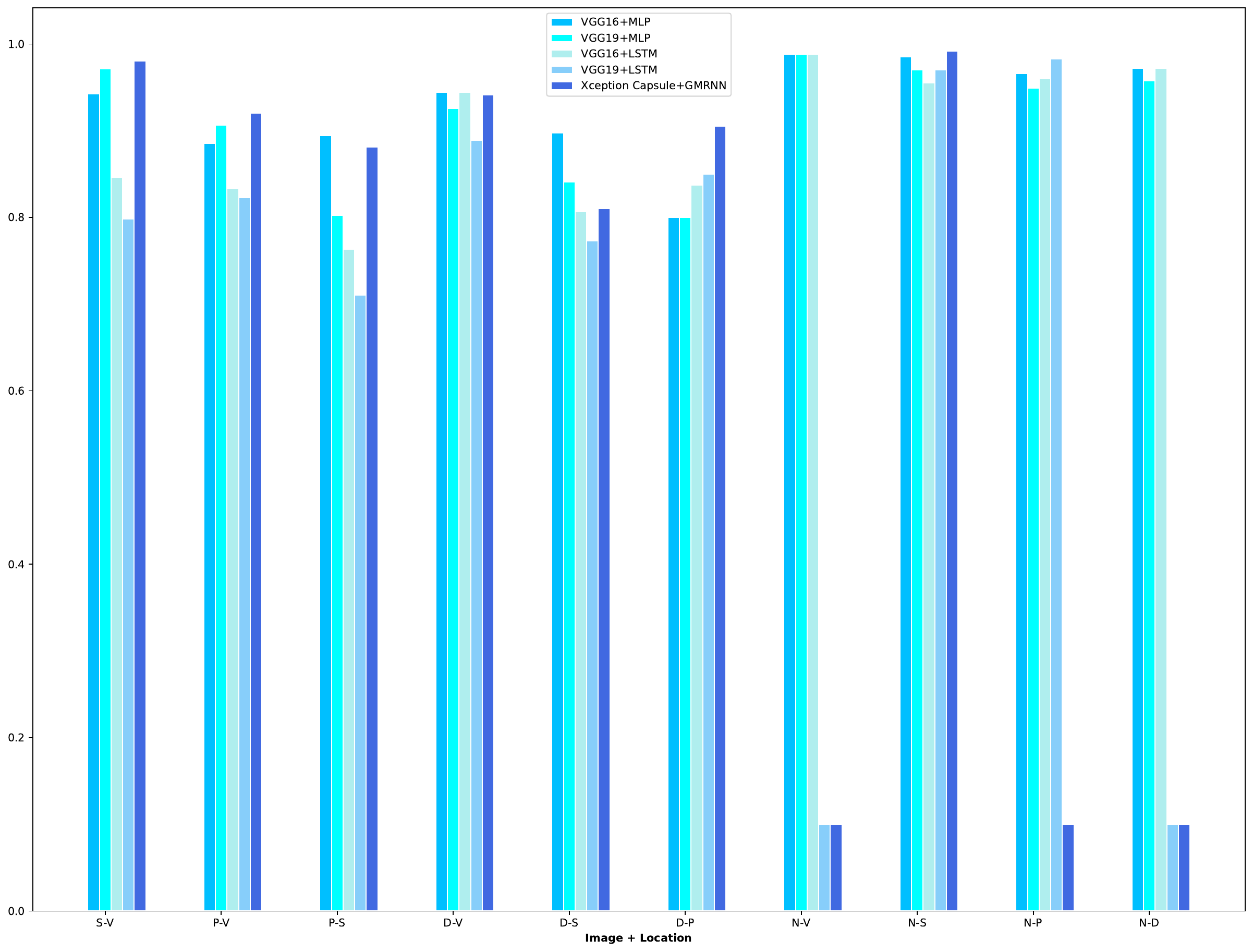}}\\
\caption{Bar plot for ten binary classifications on AZH dataset. }
\label{Fig9}
\end{figure*}

Eventually, ten binary classifications were conducted on the AZH dataset. These classifications included: (1) N vs. D, (2) N vs. P, (3) N vs. S, (4) N vs. V, (5) D vs. P, (6) D vs. S, (7) D vs. V, (8) P vs. S, (9) P vs. V, and (10) S vs. V. The highest accuracies achieved were 100\%, 100\%, 99.21\%, 100\%, 90.54\%, 81.00\%, 94.11\%, 88.12\%, 92.03\%, and 98.01\% for classifications (1), (2), (3), (4), (5), (6), (7), (8), (9), and (10) respectively. The Xception+GMRNN combination achieved the highest accuracy in all binary classifications. Detailed results are given in Table \ref{Table7}. Additionally, Figure \ref{Fig9} illustrates the bar plots for ten binary classifications on AZH dataset for further analysis.

\begin{table*}[]

\centering
\resizebox{\textwidth}{!}{
\begin{tabular}{|l|l|llllllllll|}
\hline
                                 & Model             & \multicolumn{1}{l|}{N-D}   & \multicolumn{1}{l|}{N-D}   & \multicolumn{1}{l|}{N-S}   & \multicolumn{1}{l|}{N-V}   & \multicolumn{1}{l|}{D-P}   & \multicolumn{1}{l|}{D-S}   & \multicolumn{1}{l|}{D-V}   & \multicolumn{1}{l|}{P-S}   & \multicolumn{1}{l|}{P-V}   & S-V   \\ \hline
                                 &                   & \multicolumn{10}{c|}{Accuracy}                                                                                                                                                                                                                                             \\ \hline
\multirow{3}{*}{Location}        & MLP               & \multicolumn{1}{l|}{78.87} & \multicolumn{1}{l|}{78.87} & \multicolumn{1}{l|}{74.63} & \multicolumn{1}{l|}{78.16} & \multicolumn{1}{l|}{78.75} & \multicolumn{1}{l|}{87.50} & \multicolumn{1}{l|}{89.81} & \multicolumn{1}{l|}{73.68} & \multicolumn{1}{l|}{87.50} & 93.27 \\ \cline{2-12}
                                 & LSTM              & \multicolumn{1}{l|}{77.46} & \multicolumn{1}{l|}{77.46} & \multicolumn{1}{l|}{76.12} & \multicolumn{1}{l|}{78.16} & \multicolumn{1}{l|}{78.75} & \multicolumn{1}{l|}{81.82} & \multicolumn{1}{l|}{57.41} & \multicolumn{1}{l|}{73.68} & \multicolumn{1}{l|}{85.42} & 93.27 \\ \cline{2-12}
                                 & GMRNN             & \multicolumn{1}{l|}{80.76} & \multicolumn{1}{l|}{80.76} & \multicolumn{1}{l|}{77.45} & \multicolumn{1}{l|}{78.15} & \multicolumn{1}{l|}{79.20} & \multicolumn{1}{l|}{89.12} & \multicolumn{1}{l|}{90.19} & \multicolumn{1}{l|}{74.09} & \multicolumn{1}{l|}{88.12} & 94.00 \\ \hline
\multirow{3}{*}{Image}           & VGG16             & \multicolumn{1}{l|}{98.59} & \multicolumn{1}{l|}{98.59} & \multicolumn{1}{l|}{96.61} & \multicolumn{1}{l|}{97.01} & \multicolumn{1}{l|}{81.25} & \multicolumn{1}{l|}{79.55} & \multicolumn{1}{l|}{87.96} & \multicolumn{1}{l|}{77.63} & \multicolumn{1}{l|}{84.38} & 84.62 \\ \cline{2-12}
                                 & VGG19             & \multicolumn{1}{l|}{98.59} & \multicolumn{1}{l|}{98.59} & \multicolumn{1}{l|}{97.01} & \multicolumn{1}{l|}{98.85} & \multicolumn{1}{l|}{71.25} & \multicolumn{1}{l|}{80.68} & \multicolumn{1}{l|}{87.96} & \multicolumn{1}{l|}{73.68} & \multicolumn{1}{l|}{86.46} & 86.54 \\ \cline{2-12}
                                 & Xception + Capsule          & \multicolumn{1}{l|}{99.00} & \multicolumn{1}{l|}{99.00} & \multicolumn{1}{l|}{99.01} & \multicolumn{1}{l|}{99.21} & \multicolumn{1}{l|}{90.12} & \multicolumn{1}{l|}{86.12} & \multicolumn{1}{l|}{90.43} & \multicolumn{1}{l|}{80.80} & \multicolumn{1}{l|}{86.12} & 86.32 \\ \hline
\multirow{5}{*}{Image+ Location} & VGG16 + MLP       & \multicolumn{1}{l|}{97.18} & \multicolumn{1}{l|}{97.18} & \multicolumn{1}{l|}{98.51} & \multicolumn{1}{l|}{98.85} & \multicolumn{1}{l|}{80.00} & \multicolumn{1}{l|}{89.77} & \multicolumn{1}{l|}{94.44} & \multicolumn{1}{l|}{89.47} & \multicolumn{1}{l|}{88.54} & 94.23 \\ \cline{2-12}
                                 & VGG19 + MLP       & \multicolumn{1}{l|}{95.77} & \multicolumn{1}{l|}{95.77} & \multicolumn{1}{l|}{97.01} & \multicolumn{1}{l|}{98.85} & \multicolumn{1}{l|}{80.00} & \multicolumn{1}{l|}{84.10} & \multicolumn{1}{l|}{92.59} & \multicolumn{1}{l|}{80.26} & \multicolumn{1}{l|}{90.63} & 97.12 \\ \cline{2-12}
                                 & VGG16 + LSTM      & \multicolumn{1}{l|}{97.18} & \multicolumn{1}{l|}{97.18} & \multicolumn{1}{l|}{95.52} & \multicolumn{1}{l|}{98.85} & \multicolumn{1}{l|}{83.75} & \multicolumn{1}{l|}{80.68} & \multicolumn{1}{l|}{94.44} & \multicolumn{1}{l|}{76.32} & \multicolumn{1}{l|}{83.33} & 84.62 \\ \cline{2-12}
                                 & VGG19 + LSTM      & \multicolumn{1}{l|}{100}   & \multicolumn{1}{l|}{100}   & \multicolumn{1}{l|}{97.01} & \multicolumn{1}{l|}{100}   & \multicolumn{1}{l|}{85.00} & \multicolumn{1}{l|}{77.27} & \multicolumn{1}{l|}{88.89} & \multicolumn{1}{l|}{71.05} & \multicolumn{1}{l|}{82.29} & 79.81 \\ \cline{2-12}
                                 & Xception+   GMRNN & \multicolumn{1}{l|}{100}   & \multicolumn{1}{l|}{100}   & \multicolumn{1}{l|}{99.21} & \multicolumn{1}{l|}{100}   & \multicolumn{1}{l|}{90.54} & \multicolumn{1}{l|}{81.00} & \multicolumn{1}{l|}{94.11} & \multicolumn{1}{l|}{88.12} & \multicolumn{1}{l|}{92.03} & 98.01 \\ \hline
\end{tabular}
}
\caption{Accuracy of ten binary classifications on AZH dataset. The bold represents the highest results/
accuracy achieved for each experiment.}
\label{Table7}
\end{table*}

\subsection{Sensitivity analysis}

\begin{figure*}
\centering
\subfloat[Batch size]{\includegraphics[width = 0.5\textwidth]{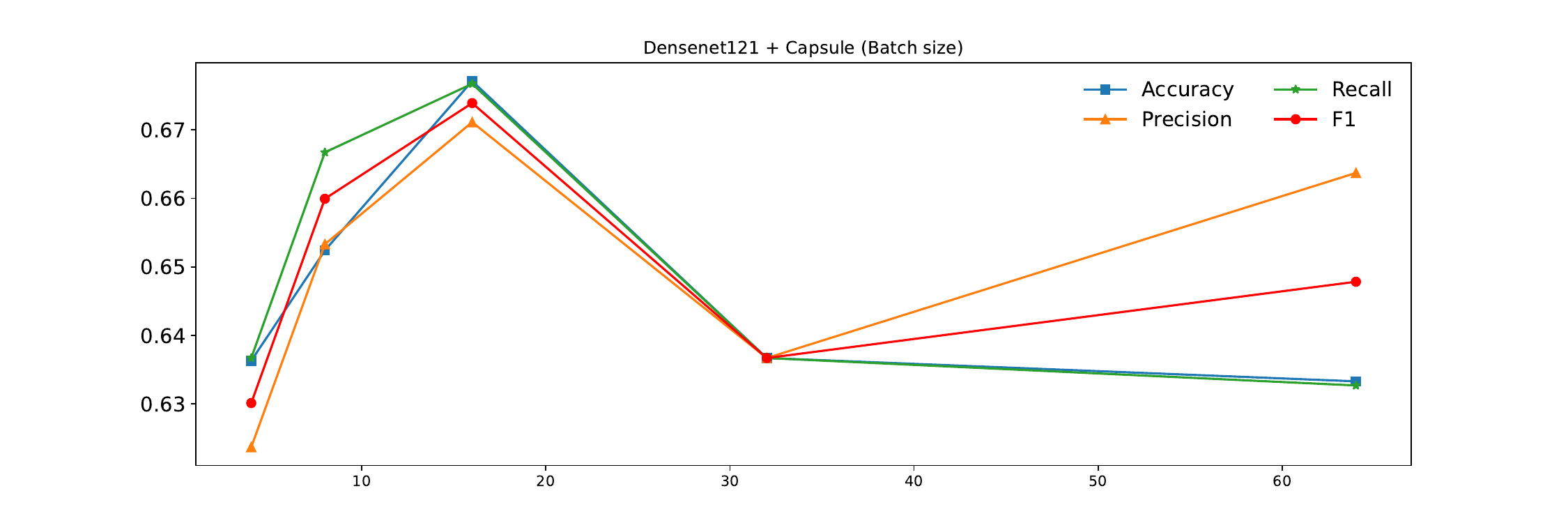}}
\subfloat[Drop size]{\includegraphics[width = 0.5\textwidth]{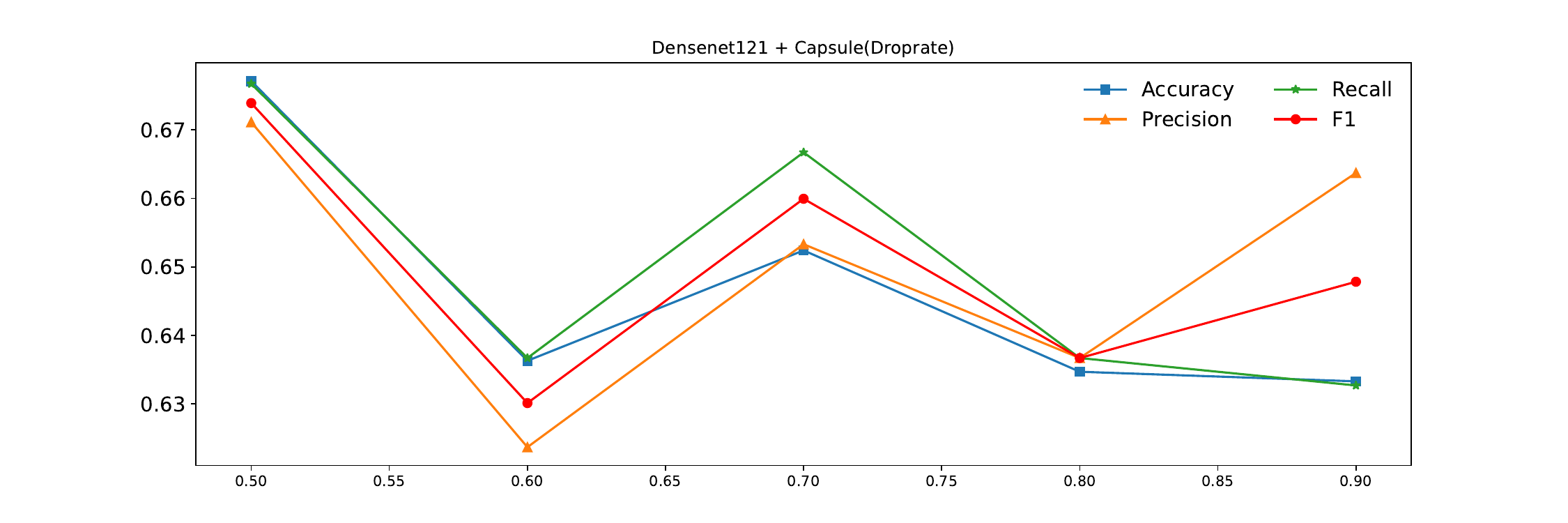}}\\
\caption{Densenet121 + Capsule hyper-parameter sensitivity analysis.}
\label{SE1}
\end{figure*}

\begin{figure*}
\subfloat[Batch size]{\includegraphics[width = 0.5\textwidth]{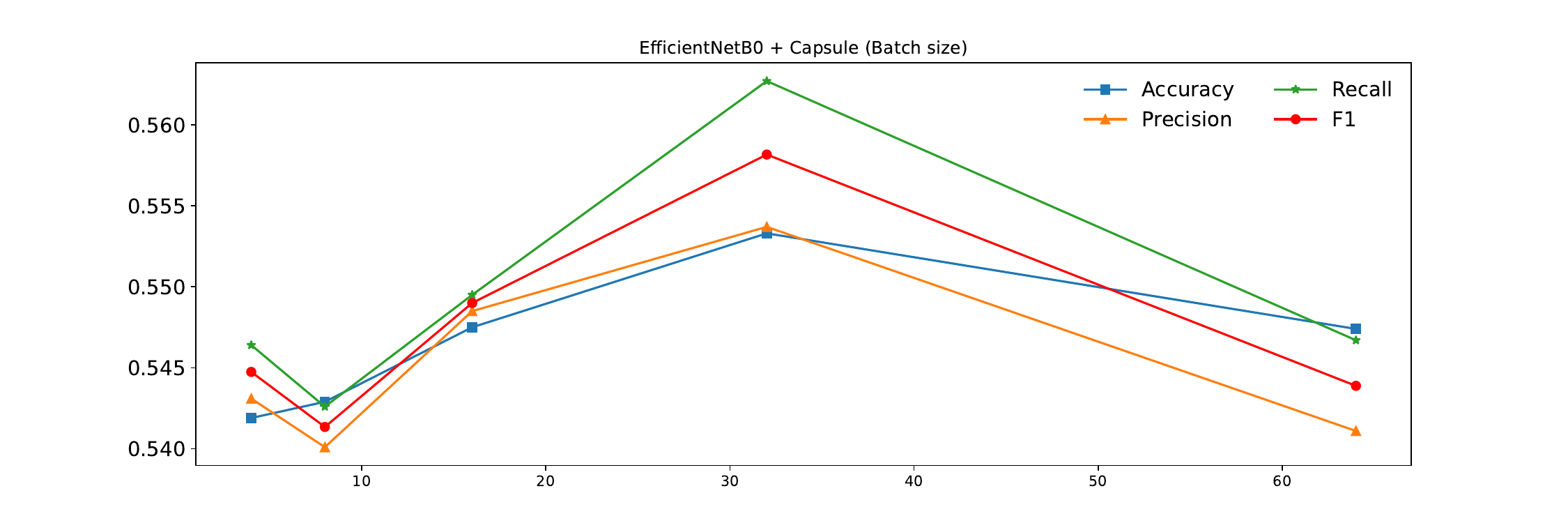}}
\subfloat[Drop size]{\includegraphics[width = 0.5\textwidth]{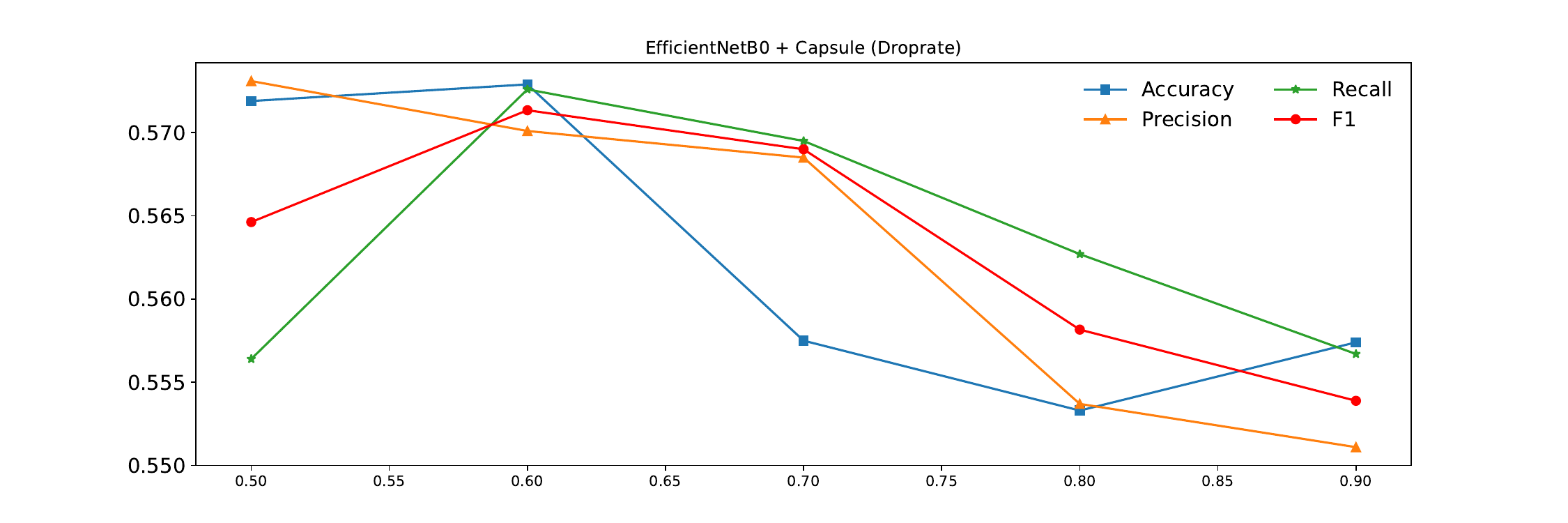}}\\
\caption{EfficientNetB0 + Capsule hyper-parameter sensitivity analysis.}
\label{SE2}
\end{figure*}

\begin{figure*}
\subfloat[Batch size]{\includegraphics[width = 0.5\textwidth]{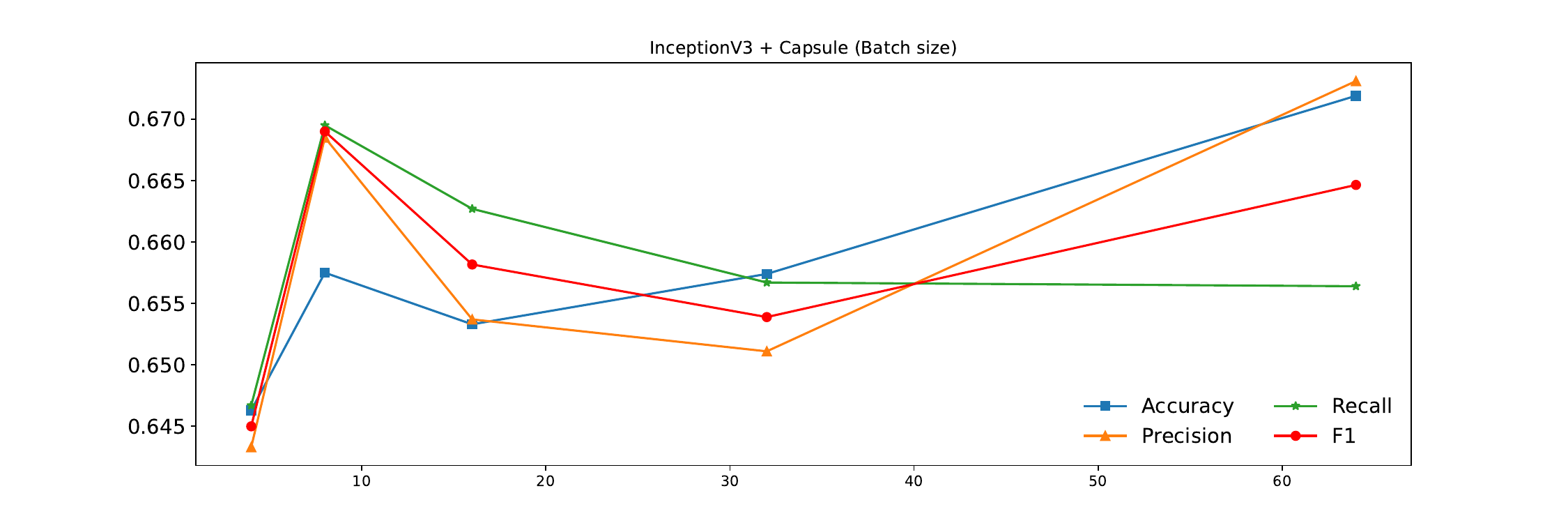}}
\subfloat[Drop size]{\includegraphics[width = 0.5\textwidth]{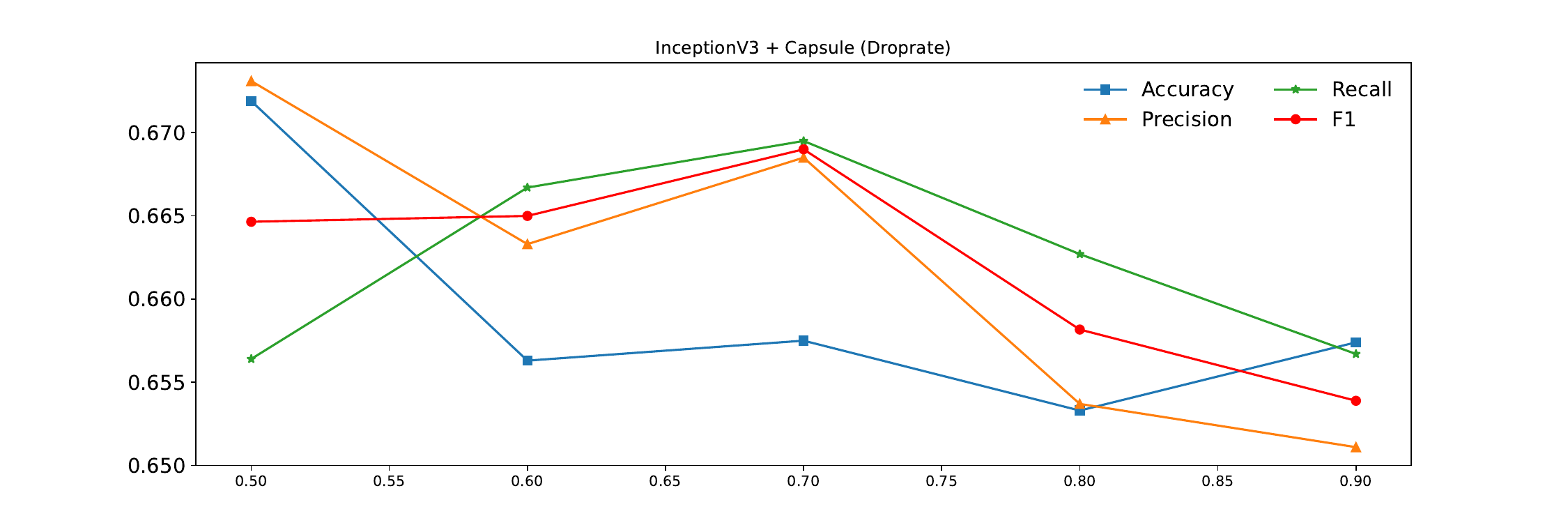}}\\
\caption{InceptionV3 + Capsule hyper-parameter sensitivity analysis.}
\label{SE3}
\end{figure*}

\begin{figure*}
\subfloat[Batch size]{\includegraphics[width = 0.5\textwidth]{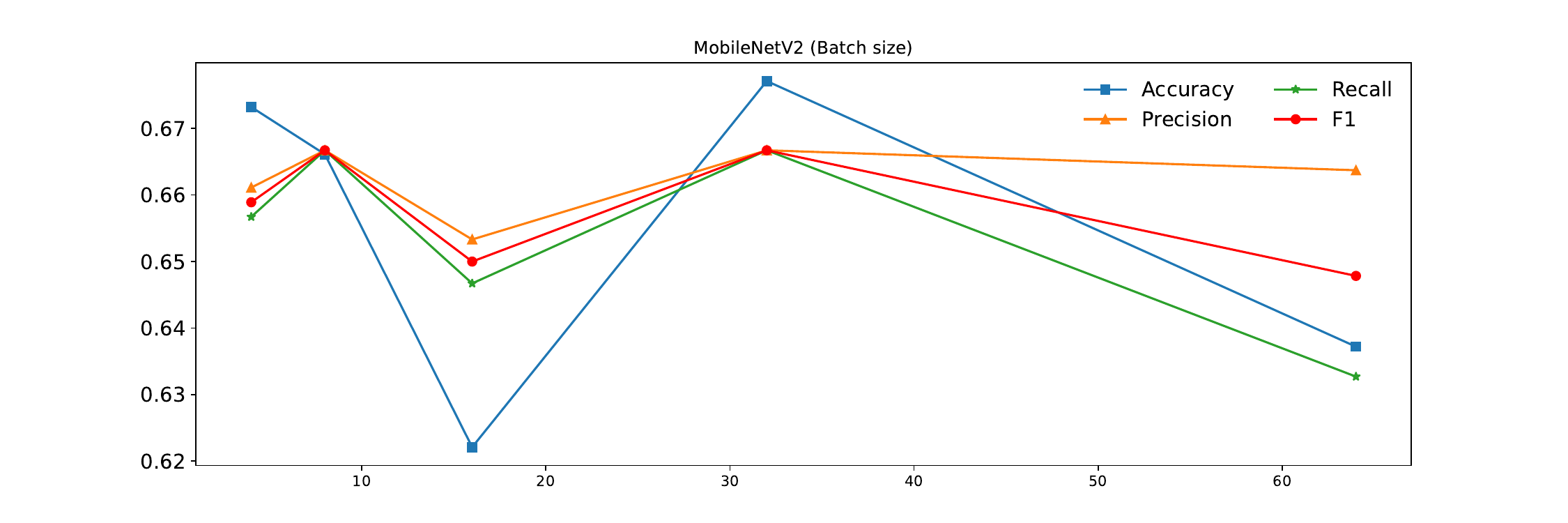}}
\subfloat[Drop size]{\includegraphics[width = 0.5\textwidth]{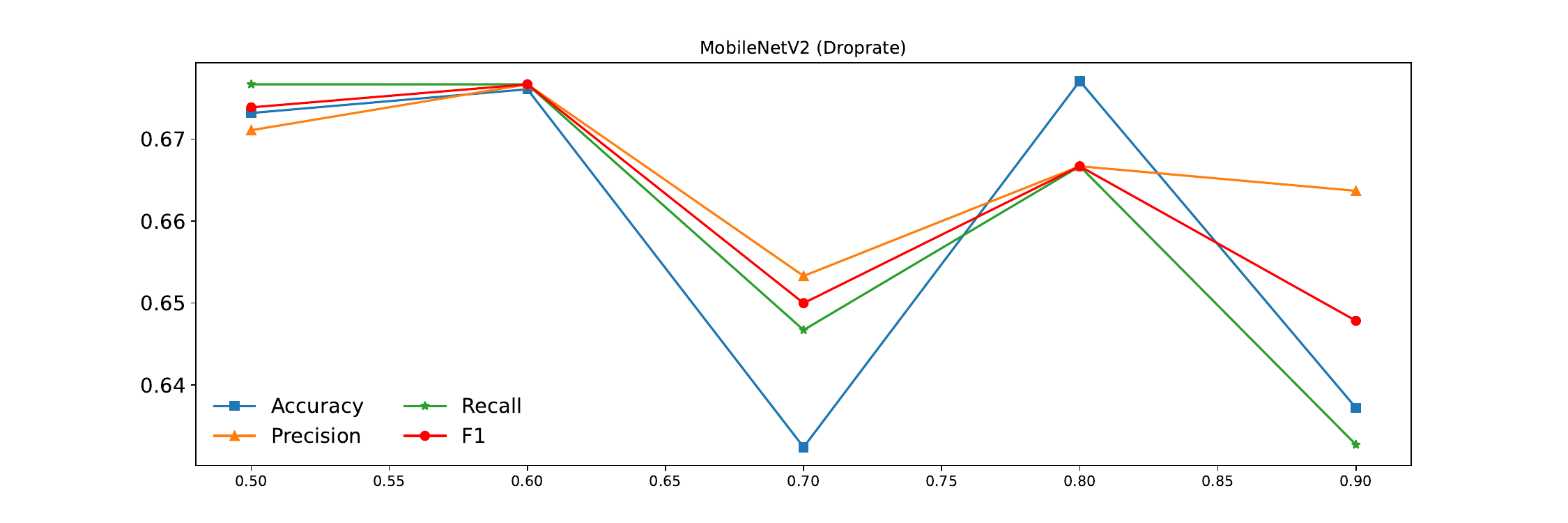}}\\
\caption{MobileNetV2 + Capsule hyper-parameter sensitivity analysis.}
\label{SE4}
\end{figure*}

\begin{figure*}
\subfloat[Batch size]{\includegraphics[width = 0.5\textwidth]{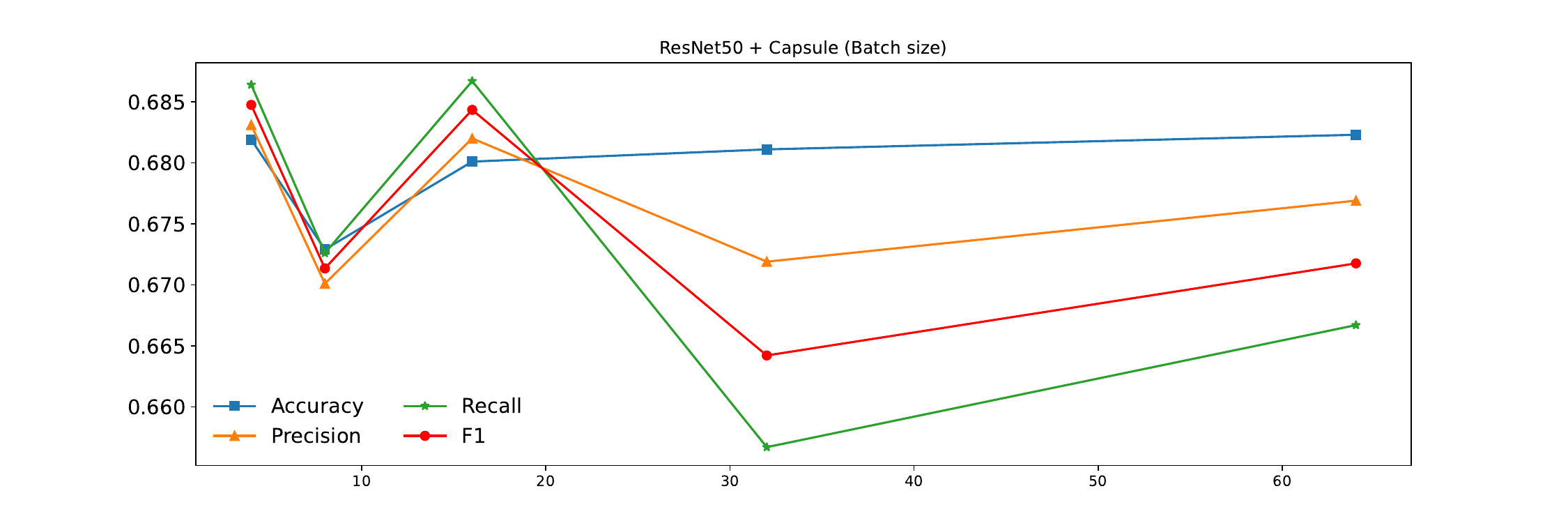}}
\subfloat[Drop size]{\includegraphics[width = 0.5\textwidth]{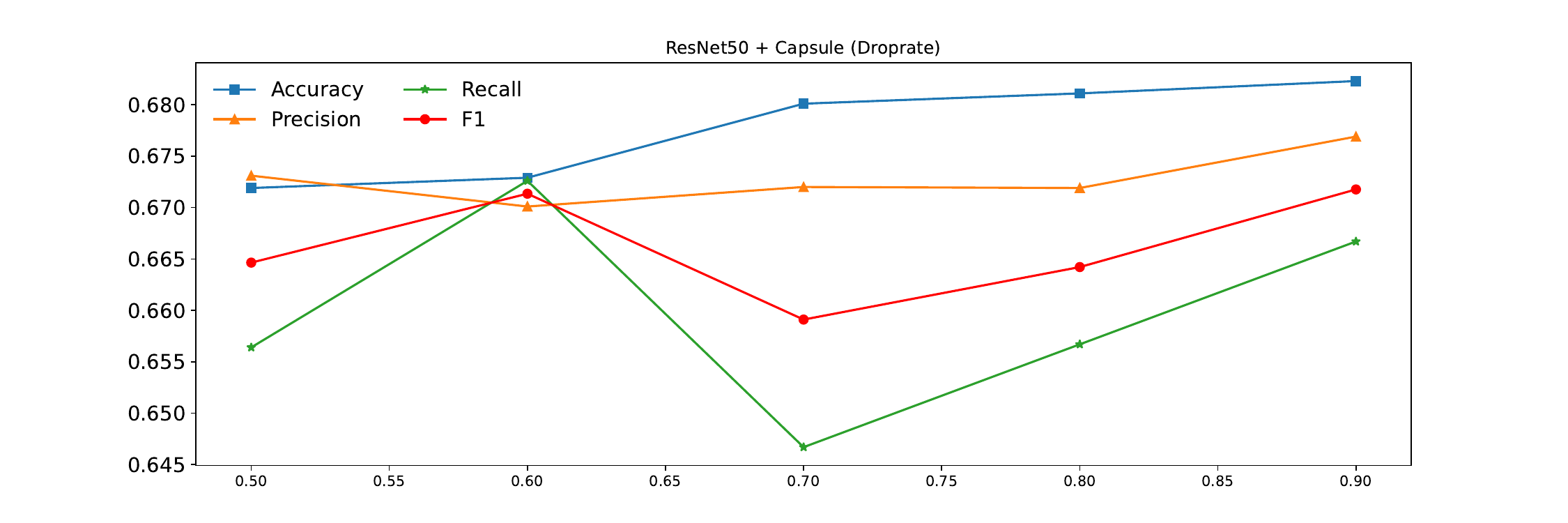}}\\
\caption{ResNet50 + Capsule hyper-parameter sensitivity analysis.}
\label{SE5}
\end{figure*}

\begin{figure*}
\subfloat[Batch size]{\includegraphics[width = 0.5\textwidth]{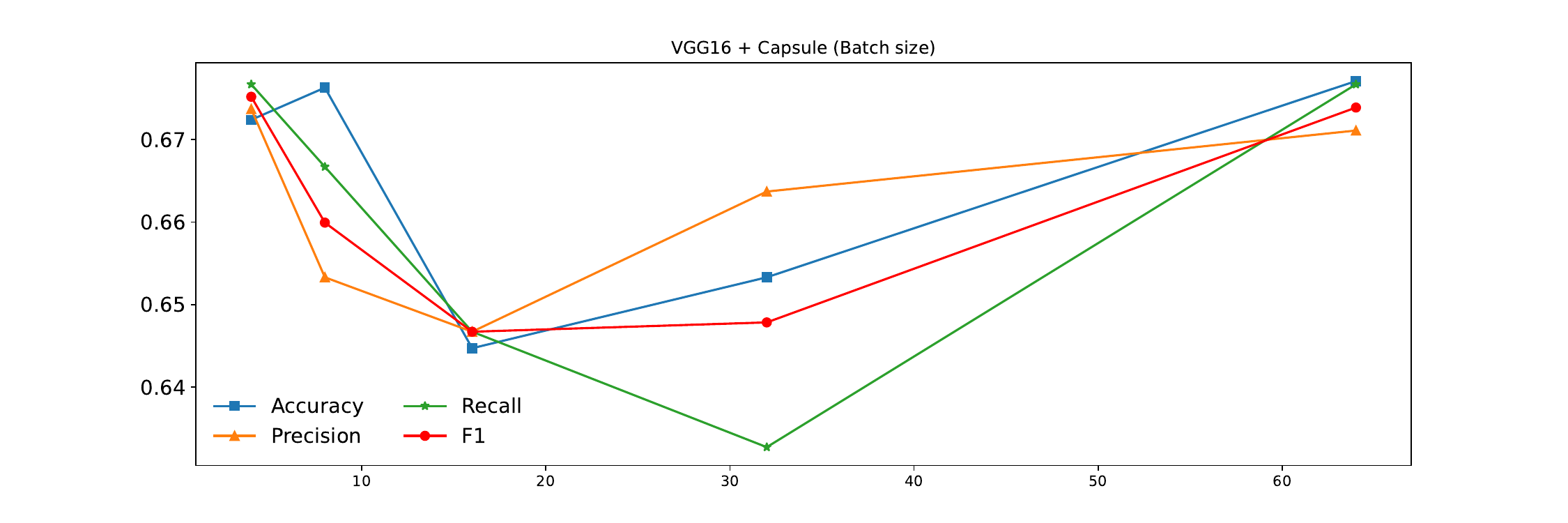}}
\subfloat[Drop size]{\includegraphics[width = 0.5\textwidth]{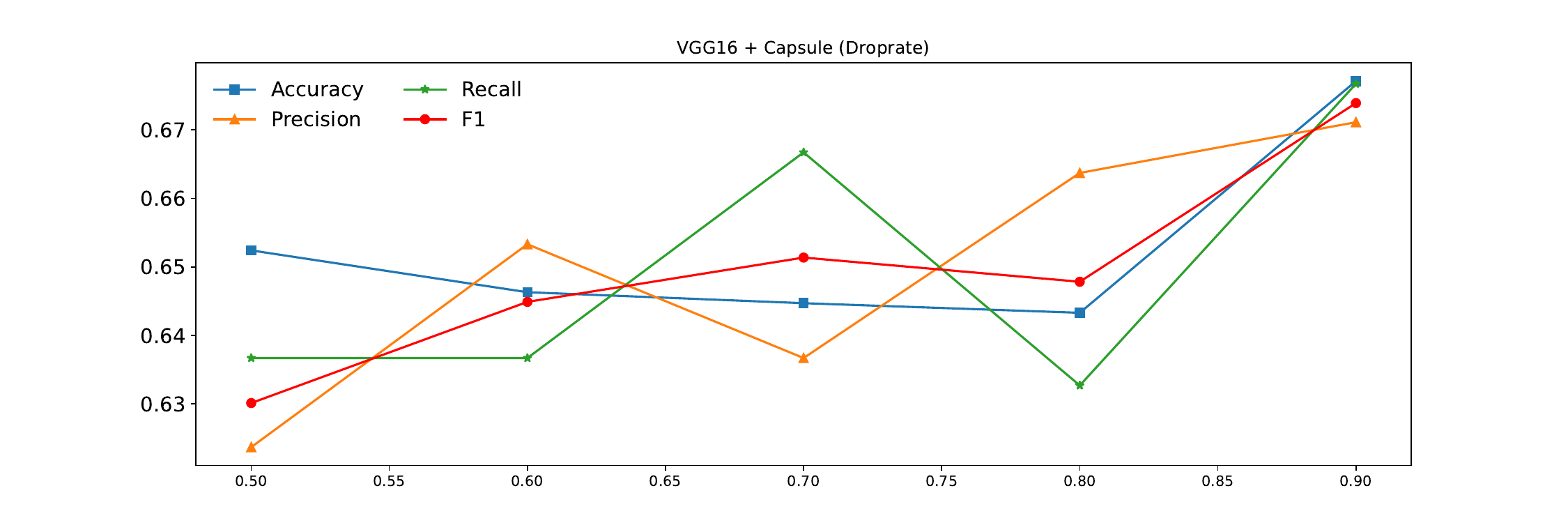}}\\
\caption{VGG16 + Capsule hyper-parameter sensitivity analysis.}
\label{SE6}
\end{figure*}

\begin{figure*}
\subfloat[Batch size]{\includegraphics[width = 0.5\textwidth]{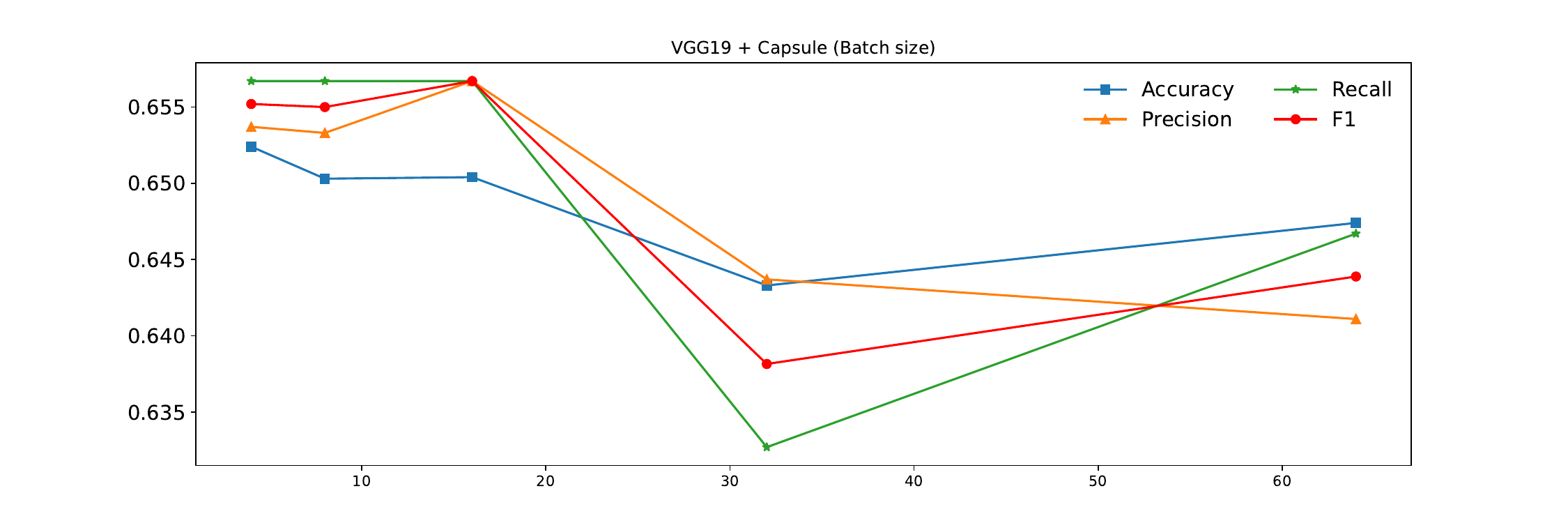}}
\subfloat[Drop size]{\includegraphics[width = 0.5\textwidth]{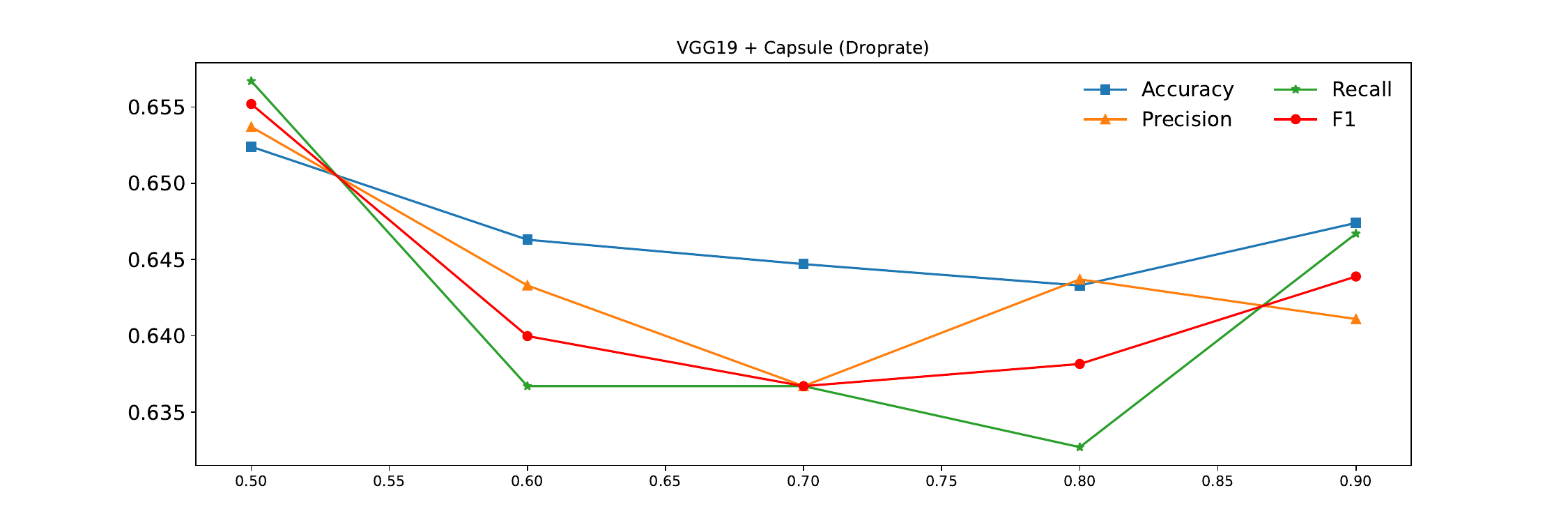}}\\
\caption{VGG16 + Capsule hyper-parameter sensitivity analysis.}
\label{SE7}
\end{figure*}

In this section, we examine the effectiveness of the proposed models' two parameters, batch size and dropout rate.
\begin{enumerate}
  \item\textbf{ Batch size:} The batch size specifies the number of samples to be published through the network (in fact, they are instantaneous network inputs). The higher the batch size, the more RAM space the program requires. Batch size is a hyper-parameter in the model and gradient descent that controls the number of training samples that must be run before updating the model's internal parameters. This parameter is the number of samples processed before updating the model. The size of a batch must be $batch_{size}\geq 1$ and $batch_{size}\preceq\sharp samples$ . Therefore, 4, 8, 16, 32, and  64 batch sizes were examined in the proposed models. The larger the batch size, the less time the training process takes. The relationship between batch size and model performance is shown in Figures \ref{SE1}, \ref{SE2}, \ref{SE3}, \ref{SE4}, \ref{SE5}, \ref{SE6}, \ref{SE7}-a.
  \item \textbf{Dropout:} Dropout randomly removes (i.e. zeroes) some neurons of a neural network during training for regularization. The idea of dropout is to force the network to learn additional representations from the input data. By randomly removing neurons, the network becomes more sensitive to the specific weights of individual neurons and more robust to noisy input data. This technique is implemented in the training phase of a neural network. During training, each neuron in the network is either retained with probability p or removed with probability 1-p. The probability p is a meta-parameter that can be adjusted . This probability was chosen between [0.5, 0.6, 0.7, 0.8, 0.9] in the proposed models. According to the results of Figures \ref{SE1}, \ref{SE2}, \ref{SE3}, \ref{SE4}, \ref{SE5}, \ref{SE6}, \ref{SE7}-b, the best dropout rate in the proposed models is 0.5.
\end{enumerate}

\begin{figure*}
\subfloat[Batch size]{\includegraphics[width = 0.5\textwidth]{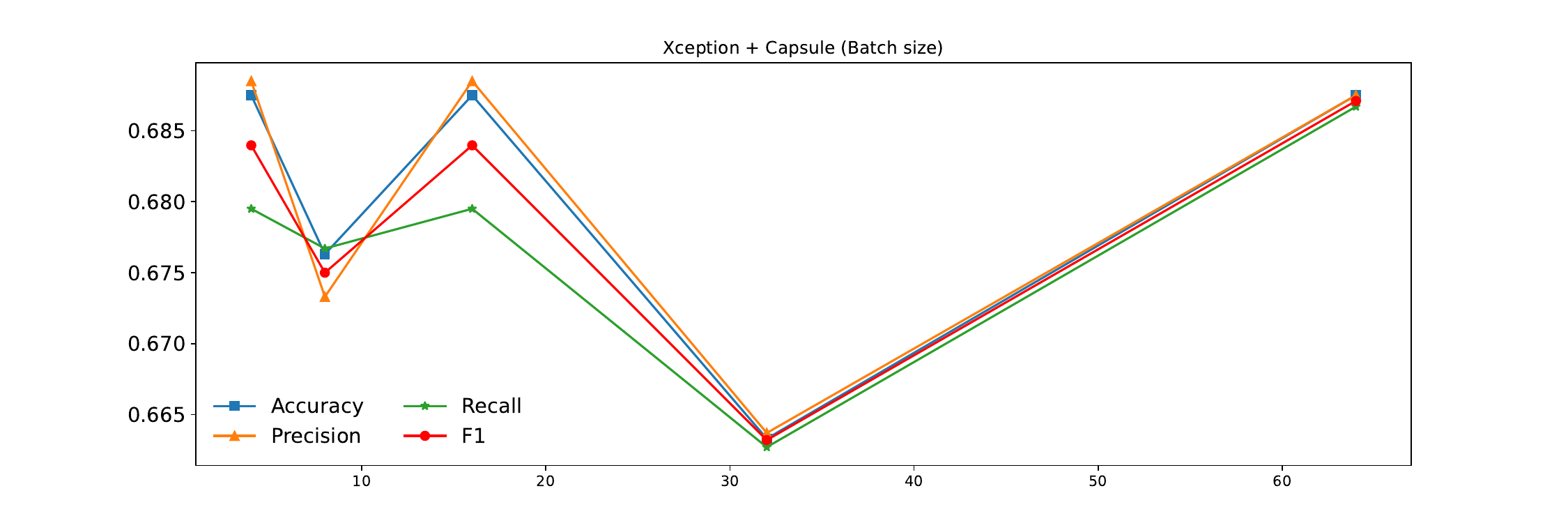}}
\subfloat[Drop size]{\includegraphics[width = 0.5\textwidth]{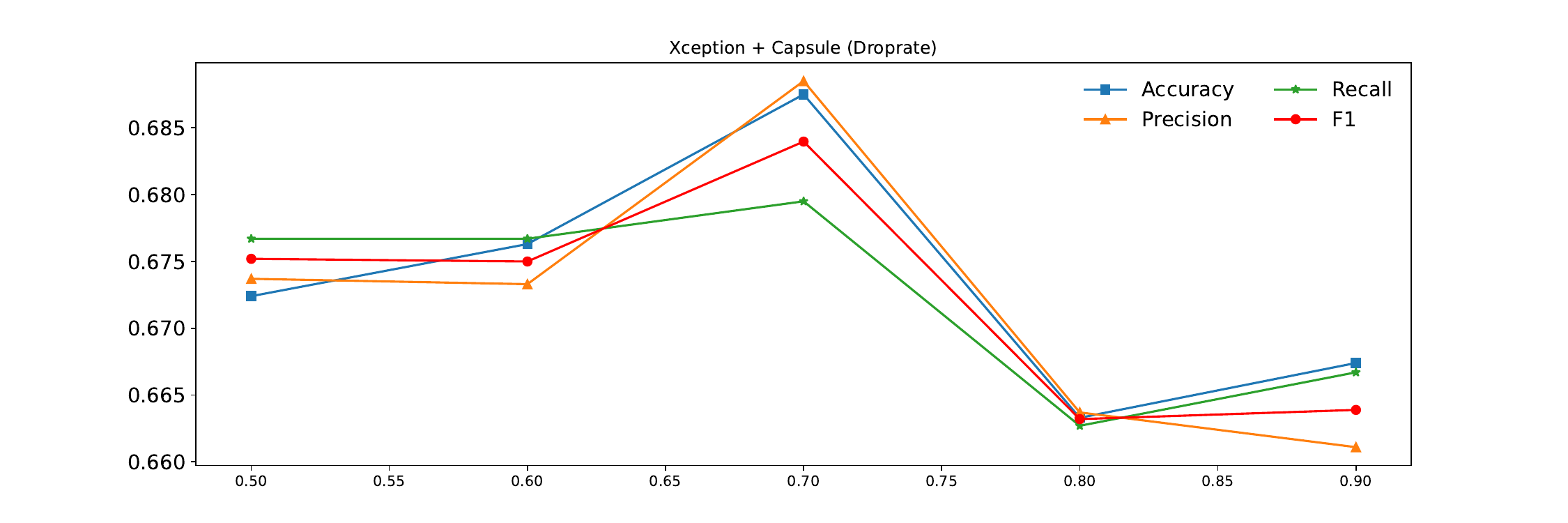}}\\
\caption{Xception + Capsule hyper-parameter sensitivity analysis.}
\label{SE8}
\end{figure*}

\section{Conclusion}
The main goal of the article was to present a wound multimodal classification (WMC) approach using wound images and their corresponding locations. Deep learning structures have achieved great results in classification, but the main problem of these models is random weighting, which leads to different results in different execution rounds. For this purpose, transfer learning based on Xception and Image-net weights were used. Also, the capsule network was placed at the end of the output layer of Xception to maintain the relationships between features, so that Xception can be used for feature extraction and the capsule can be used to learn features. A basic step in the proposed approach is to provide the GMRNN gate. This gate uses Gaussian distributions of locations to learn locations related to wounds. This approach was able to obtain more acceptable results than other existing approaches in different input and output modes. Accurate classification of wound types can help doctors diagnose wound problems more quickly and find appropriate treatment plans. A large number of experiments were conducted with a wide range of binary, 3-class, 4-class, 5-class, and 6-class classifications on three datasets. The results produced by the polynomial network were much better than the results produced by the single input, and these results beat all previous experimental results. In the future, the goal is to use GAN for data augmentation. Also, due to the fact that the data is unbalanced, cost-sensitive functions such as \cite{TT1} can be used. Based on our investigation, Twins\cite{TT2}  is even able to provide better results.

\end{document}